\documentclass[10pt,twocolumn,letterpaper]{article}

\usepackage{cvpr}
\makeatletter
\@namedef{ver@everyshi.sty}{}
\makeatother
\usepackage{times}
\usepackage{epsfig}
\usepackage{graphicx}
\usepackage{amsmath}
\usepackage{amssymb}
\usepackage{xcolor}
\usepackage{caption}
\usepackage{multirow}
\usepackage{subcaption}
\usepackage[numbers,sort,compress]{natbib}
\usepackage{booktabs}
\usepackage{makecell}
\usepackage{pifont}
\usepackage{csquotes}
\usepackage{bm}

\usepackage{booktabs}
\usepackage{changes}
\usepackage{enumitem}
\usepackage{soul}
\usepackage{multirow}
\definechangesauthor[color=blue, name={Zijun Wei}]{ZJ}
\usepackage[pagebackref=true,breaklinks=true,letterpaper=true,colorlinks,bookmarks=false]{hyperref}

\cvprfinalcopy 

\def\cvprPaperID{7339} 

\begin{document}
\def\mA{\mathcal{A}}
\def\mB{\mathcal{B}}
\def\mC{\mathcal{C}}
\def\mD{\mathcal{D}}
\def\mE{\mathcal{E}}
\def\mF{\mathcal{F}}
\def\mG{\mathcal{G}}
\def\mH{\mathcal{H}}
\def\mI{\mathcal{I}}
\def\mJ{\mathcal{J}}
\def\mK{\mathcal{K}}
\def\mL{\mathcal{L}}
\def\mM{\mathcal{M}}
\def\mN{\mathcal{N}}
\def\mO{\mathcal{O}}
\def\mP{\mathcal{P}}
\def\mQ{\mathcal{Q}}
\def\mR{\mathcal{R}}
\def\mS{\mathcal{S}}
\def\mT{\mathcal{T}}
\def\mU{\mathcal{U}}
\def\mV{\mathcal{V}}
\def\mW{\mathcal{W}}
\def\mX{\mathcal{X}}
\def\mY{\mathcal{Y}}
\def\mZ{\mathcal{Z}}

\def\1n{\mathbf{1}_n}
\def\0{\mathbf{0}}
\def\1{\mathbf{1}}

\def\A{{\bf A}}
\def\B{{\bf B}}
\def\C{{\bf C}}
\def\D{{\bf D}}
\def\E{{\bf E}}
\def\F{{\bf F}}
\def\G{{\bf G}}
\def\H{{\bf H}}
\def\I{{\bf I}}
\def\J{{\bf J}}
\def\K{{\bf K}}
\def\L{{\bf L}}
\def\M{{\bf M}}
\def\N{{\bf N}}
\def\O{{\bf O}}
\def\P{{\bf P}}
\def\Q{{\bf Q}}
\def\R{{\bf R}}
\def\S{{\bf S}}
\def\T{{\bf T}}
\def\U{{\bf U}}
\def\V{{\bf V}}
\def\W{{\bf W}}
\def\X{{\bf X}}
\def\Y{{\bf Y}}
\def\Z{{\bf Z}}

\def\a{{\bf a}}
\def\b{{\bf b}}
\def\c{{\bf c}}
\def\d{{\bf d}}
\def\e{{\bf e}}
\def\f{{\bf f}}
\def\g{{\bf g}}
\def\h{{\bf h}}
\def\i{{\bf i}}
\def\j{{\bf j}}
\def\k{{\bf k}}
\def\l{{\bf l}}
\def\m{{\bf m}}
\def\n{{\bf n}}
\def\o{{\bf o}}
\def\p{{\bf p}}
\def\q{{\bf q}}
\def\r{{\bf r}}
\def\s{{\bf s}}
\def\t{{\bf t}}
\def\u{{\bf u}}
\def\v{{\bf v}}
\def\w{{\bf w}}
\def\x{{\bf x}}
\def\y{{\bf y}}
\def\z{{\bf z}}

\def\balpha{\mbox{\boldmath{$\alpha$}}}
\def\bbeta{\mbox{\boldmath{$\beta$}}}
\def\bdelta{\mbox{\boldmath{$\delta$}}}
\def\bgamma{\mbox{\boldmath{$\gamma$}}}
\def\blambda{\mbox{\boldmath{$\lambda$}}}
\def\bsigma{\mbox{\boldmath{$\sigma$}}}
\def\btheta{\mbox{\boldmath{$\theta$}}}
\def\bomega{\mbox{\boldmath{$\omega$}}}
\def\bxi{\mbox{\boldmath{$\xi$}}}
\def\bnu{\mbox{\boldmath{$\nu$}}}                                  
\def\bphi{\mbox{\boldmath{$\phi$}}}
\def\bmu{\mbox{\boldmath{$\mu$}}}

\def\bDelta{\mbox{\boldmath{$\Delta$}}}
\def\bOmega{\mbox{\boldmath{$\Omega$}}}
\def\bPhi{\mbox{\boldmath{$\Phi$}}}
\def\bLambda{\mbox{\boldmath{$\Lambda$}}}
\def\bSigma{\mbox{\boldmath{$\Sigma$}}}
\def\bGamma{\mbox{\boldmath{$\Gamma$}}}

\newcommand{\myminimum}[1]{\mathop{\textrm{minimum}}_{#1}}
\newcommand{\mymaximum}[1]{\mathop{\textrm{maximum}}_{#1}}    
\newcommand{\mymin}[1]{\mathop{\textrm{minimize}}_{#1}}
\newcommand{\mymax}[1]{\mathop{\textrm{maximize}}_{#1}}
\newcommand{\mymins}[1]{\mathop{\textrm{min.}}_{#1}}
\newcommand{\mymaxs}[1]{\mathop{\textrm{max.}}_{#1}}  
\newcommand{\myargmin}[1]{\mathop{\textrm{argmin}}_{#1}} 
\newcommand{\myargmax}[1]{\mathop{\textrm{argmax}}_{#1}} 
\newcommand{\myst}{\textrm{s.t. }}

\newcommand{\denselist}{\itemsep -1pt}
\newcommand{\sparselist}{\itemsep 1pt}

\definecolor{pink}{rgb}{0.9,0.5,0.5}
\definecolor{purple}{rgb}{0.5, 0.4, 0.8}   
\definecolor{gray}{rgb}{0.3, 0.3, 0.3}
\definecolor{mygreen}{rgb}{0.2, 0.6, 0.2}

\newcommand{\cyan}[1]{\textcolor{cyan}{#1}}
\newcommand{\red}[1]{\textcolor{red}{#1}}  
\newcommand{\blue}[1]{\textcolor{blue}{#1}}
\newcommand{\magenta}[1]{\textcolor{magenta}{#1}}
\newcommand{\pink}[1]{\textcolor{pink}{#1}}
\newcommand{\green}[1]{\textcolor{green}{#1}} 
\newcommand{\gray}[1]{\textcolor{gray}{#1}}    
\newcommand{\mygreen}[1]{\textcolor{mygreen}{#1}}    
\newcommand{\purple}[1]{\textcolor{purple}{#1}}       

\definecolor{greena}{rgb}{0.4, 0.5, 0.1}
\newcommand{\greena}[1]{\textcolor{greena}{#1}}

\definecolor{bluea}{rgb}{0, 0.4, 0.6}
\newcommand{\bluea}[1]{\textcolor{bluea}{#1}}
\definecolor{reda}{rgb}{0.6, 0.2, 0.1}
\newcommand{\reda}[1]{\textcolor{reda}{#1}}

\def\changemargin#1#2{\list{}{\rightmargin#2\leftmargin#1}\item[]}
\let\endchangemargin=\endlist
                                               
\newcommand{\cm}[1]{}

\newcommand{\mtodo}[1]{{\color{red}$\blacksquare$\textbf{[TODO: #1]}}}
\newcommand{\myheading}[1]{\vspace{1ex}\noindent \textbf{#1}}
\newcommand{\htimesw}[2]{\mbox{$#1$$\times$$#2$}}


\newif\ifshowsolution
\showsolutiontrue

\ifshowsolution  
\newcommand{\Comment}[1]{\paragraph{\bf $\bigstar $ COMMENT:} {\sf #1} \bigskip}
\newcommand{\Solution}[2]{\paragraph{\bf $\bigstar $ SOLUTION:} {\sf #2} }
\newcommand{\Mistake}[2]{\paragraph{\bf $\blacksquare$ COMMON MISTAKE #1:} {\sf #2} \bigskip}
\else
\newcommand{\Solution}[2]{\vspace{#1}}
\fi

\newcolumntype{L}[1]{>{\raggedright\let\newline\\\arraybackslash\hspace{0pt}}m{#1}}
\newcolumntype{C}[1]{>{\centering\let\newline\\\arraybackslash\hspace{0pt}}m{#1}}
\newcolumntype{R}[1]{>{\raggedleft\let\newline\\\arraybackslash\hspace{0pt}}m{#1}}

\newcommand{\truefalse}{
\begin{enumerate}
	\item True
	\item False
\end{enumerate}
}

\newcommand{\yesno}{
\begin{enumerate}
	\item Yes
	\item No
\end{enumerate}
}

\newcommand{\Sref}[1]{Sec.~\ref{#1}}
\newcommand{\Eref}[1]{Eq.~(\ref{#1})}
\newcommand{\Fref}[1]{Fig.~\ref{#1}}
\newcommand{\Tref}[1]{Tab.~\ref{#1}}

\newcommand{\mh}[1]{\textcolor{red}{[Minh: {#1}]}}
\newcommand{\lh}[1]{\textcolor{blue}{[Lihan: {#1}]}}
\makeatletter
\def\blfootnote{\gdef\@thefnmark{}\@footnotetext}
\makeatother
\title{Predicting Goal-directed Human Attention Using \\ Inverse Reinforcement Learning}

\author{
Zhibo Yang\textsuperscript{1*}, Lihan Huang\textsuperscript{1*}, Yupei Chen\textsuperscript{1}, Zijun Wei\textsuperscript{2}, Seoyoung Ahn\textsuperscript{1}, \\  Gregory Zelinsky\textsuperscript{1}, Dimitris Samaras\textsuperscript{1}, Minh Hoai\textsuperscript{1} \\
\textsuperscript{1}Stony Brook University,~~~  \textsuperscript{2}Adobe Inc.
}

\maketitle

\begin{abstract}
   Being able to predict human gaze behavior has obvious importance for behavioral vision and for computer vision applications. Most models have mainly focused on predicting free-viewing behavior using saliency maps, but these predictions do not generalize to goal-directed behavior, such as when a person searches for a visual target object. We propose the first inverse reinforcement learning (IRL) model to learn the internal reward function and policy used by humans during visual search. The viewer's internal belief states were modeled as dynamic contextual belief maps of object locations. These maps were learned by IRL and then used to predict behavioral scanpaths for multiple target categories. To train and evaluate our IRL model we created COCO-Search18, which is now the largest dataset of high-quality search fixations in existence. COCO-Search18 has 10 participants searching for each of 18 target-object categories in 6202 images, making about 300,000 goal-directed fixations. When trained and evaluated on COCO-Search18, the IRL model outperformed baseline models in predicting search fixation scanpaths, both in terms of similarity to human search behavior and search efficiency. Finally, reward maps recovered by the IRL model reveal distinctive target-dependent patterns of object prioritization, which we interpret as a learned object context. 
   \blfootnote{Code and dataset are available at \url{https://github.com/cvlab-stonybrook/Scanpath_Prediction}.}
   \blfootnote{*Equal contribution.}
   
  
\end{abstract}

\section{Introduction}

\begin{figure}[t]
  \centering
  \includegraphics[width=1.0\linewidth]{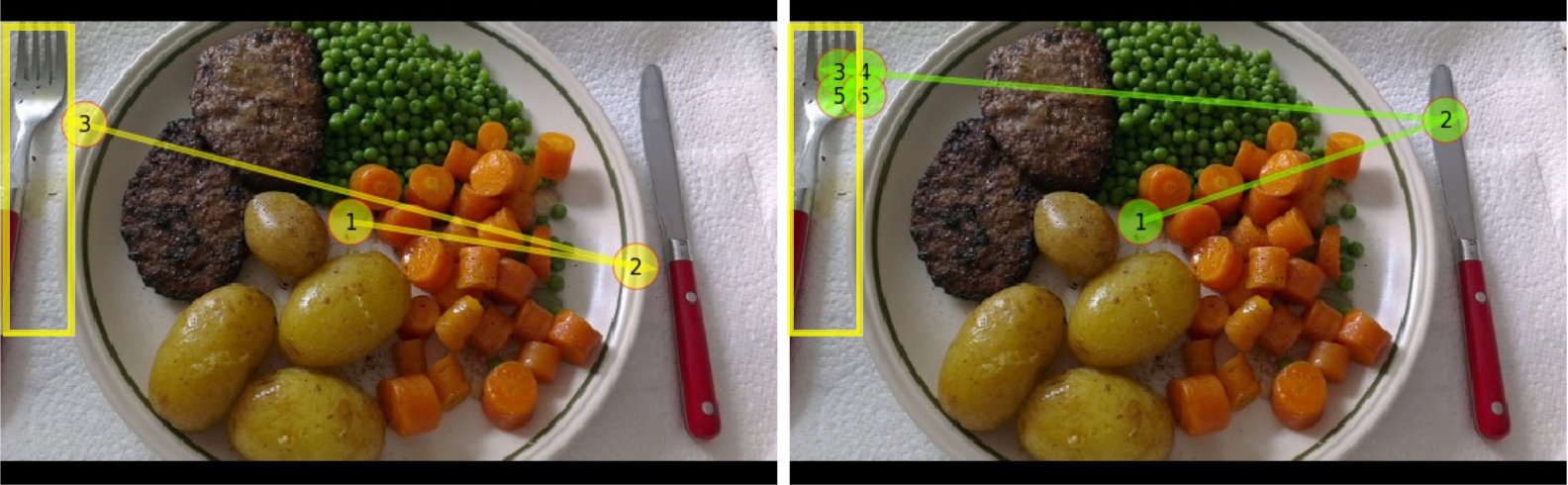}
  \vskip -0.1in
  \caption{{\bf Predicting fixations in a visual search task}. Left: behavioral scanpath shown in yellow. Right: predicted scanpath in green. The search target is the fork, shown in the yellow bounding box. }
  \label{fig:teaser}
  \vskip -0.2in
\end{figure}

Human visual attention comes in two forms. One is bottom-up, where prioritization is based solely on processing of the visual input. The other is top-down, where prioritization is based on a myriad of top-down information sources (the object context of a scene, the semantic relationships between objects, etc.~\cite{wolfe2017five,eckstein2011visual,nakayama2011situating}). When your food arrives at a restaurant, among your very first movements of attention will likely be to the fork and knife (\Fref{fig:teaser}), because these objects are important to your goal of having dinner. Goal-directed attention control underlies all the tasks that we {\bf try} to do, thereby making its prediction a vastly more challenging and important problem than predicting the bottom-up control of attention by a visual input. Perhaps the strongest form of top-down attention control is in the definition of a target goal, with visual search being arguably the simplest goal-directed tasks---there is a target goal and the task is to find it. Humans are highly efficient and flexible in the image locations that they choose to fixate while searching for a target-object goal, making the prediction of human search behavior important for both behavioral and computer vision, and in particular for robotic visual systems such as robotic search~\cite{elder2006attentive, narayanan2016perch}.
In this paper, we introduce Inverse Reinforcement Learning as a computational model of human attention in visual search.

\myheading{Gaze prediction in visual search}.
The aim of gaze prediction is to predict the patterns of fixations made by people during image viewing. These patterns can be either spatial (fixation density maps) or spatial+temporal (scanpaths). Most fixation-prediction models are in the context of a free-viewing task, where the prioritization of locations for fixation is believed to be controlled by bottom-up saliency. Since the seminal work by \citet{IttiPAMI98}, the use of saliency maps to predict free-viewing fixations has grown into a sizeable literature \cite{borji2013state, kummerer2017understanding, kruthiventi2017deepfix, huang2015salicon, kummerer2014deep, cornia2018predicting,jiang2015salicon,jetley2016end, borji2015salient, masciocchi2009everyone, berg2009free}. However, the predictions of saliency models do not generalize to fixations made in goal-directed attention tasks, such as the search for a target object~\cite{henderson2007visual, koehler2014saliency}. 

A critical difference between a search task and a free-viewing task is that search fixations are guided towards a target-object goal, whereas in free viewing there are no explicit goals. Target guidance during search was first quantified using simple targets having simple features that were known to the searcher~\cite{wolfe1994guided}. This work was followed by computational models that used images of objects and scenes as inputs~\cite{ehinger2009modelling, zelinsky2008theory, zelinsky2013modeling}, which further expanded to include target spatial relationships~\cite{aydemir2011search} and global scene context~\cite{torralba2006contextual}. Target spatial relationships and scene context are important factors that substantially affect human search efficiency. When searching for objects in scenes, scene context can guide attention to the target. To date, there have been very few deep network models attempting to predict human search fixations \cite{zhang2018finding, adeli2018deep,Wei-etal-NIPS16}. What all of these models have in common, however, is that they use some algorithm, and knowledge about a particular source of information (target features, meaning, context, etc), to prioritize image locations for fixation selection.

\myheading{Inverse Reinforcement Learning}. Our approach to search-fixation prediction is exactly the opposite. Rather than using an algorithm to prioritize locations in an image, here we use Inverse Reinforcement Learning (IRL)~\cite{Ng-Russell-ICML00, Abbeel-etal-ICML04,Ziebart-etal-AAAI08,ho2016generative,fu2017learning} to learn sequences of search fixations by treating each as a potential source of reward. IRL, a form of imitation learning, refers to the recovery of an expert's underlying reward function through repeated observation. Most IRL algorithms \cite{Ziebart-etal-AAAI08,fu2017learning,wulfmeier2015maximum} simultaneously learn an optimal policy and the reward function on which the policy is optimized. Although early IRL algorithms \cite{Ng-Russell-ICML00,Ziebart-etal-AAAI08} were often restricted to problems having low-dimensional state spaces, \citet{wulfmeier2015maximum} proposed deep maximum entropy IRL to handle raw image inputs. Recent work~\cite{ho2016generative,fu2017learning} applies adversarial training \cite{Goodfellow-etal-NIPS14} to learn the underlying reward function and the policy, treating each as (part of) the discriminator and the generator in adversarial training, respectively. The discriminator assigns high reward to an expert's behavior and low reward to a non-expert's behavior, where behavior is represented as state-action pairs. The generator/policy is optimized using a reinforcement learning algorithm to get higher reward by behaving more like the expert. In this work, we use the GAIL (generative adversarial imitation learning) algorithm~\cite{ho2016generative}, given its ability to imitate behaviors in complex and high-dimensional environments \cite{ho2016generative}. Using this approach, a unified information-maximization framework is defined for combining diverse sources of information for the purpose of selecting maximally-rewarding locations to fixate, thus increasing both the accuracy and applicability of predicting human search fixations. 

\begin{table}[]
\begin{center}
\setlength{\tabcolsep}{2pt}
\begin{tabular}{lcrrrr}
\toprule 
Dataset & Search & Image & Class & Subj/img & Fixation\\
\midrule 
SALICON \cite{jiang2015salicon} & \ding{55} & 10000  & - & 60 & 4600K* \\
POET \cite{papadopoulos2014training} & \ding{55} & 6270 & 10 & 5 & 178K \\
People900 \cite{ehinger2009modelling} & \ding{51} & 912 & 1 & 14 & 55K \\
MCS \cite{zelinsky2019benchmarking} & \ding{51} & 2183 & 2 & 1-4 & 16K  \\
PET \cite{gilani2015pet} & \ding{51} & 4135 & 6 & 4 & 30K \\
COCO-Search18 & \ding{51} & 6202 & 18 & 10 & 300K \\
\bottomrule 
\end{tabular}
\vskip -0.1in
\caption{{\bf Comparison of fixation datasets}. Previous datasets either did not use a search task, or had very few target-object classes, subjects, or fixations. *: Fixations are approximated by mouse clicks.
}
\label{table: dataset cmp}
\end{center}
\vskip -0.3in
\end{table}

\myheading{Search Fixation Datasets}. Another significant contribution of our work is the introduction of the COCO-Search18 dataset, which is currently the world's largest dataset of images that have been annotated with the gaze fixations made by humans during search. COCO-Search18 is needed because the best models of goal-directed attention will likely be models trained on goal-directed behavior. Using free-viewing as an example, the currently best model of free-viewing fixations is DeepGaze II \cite{kummerer2017understanding}, which is a deep network trained on SALICON \cite{jiang2015salicon}. SALICON is a crowd-sourced dataset consisting of images that were annotated with human mouse clicks indicating attentionally salient image locations. Without SALICON, DeepGaze II and models like it, would not have been possible, and our understanding of free-viewing behavior, widely believed to reflect bottom-up attention control, would be greatly diminished.

There is nothing comparable to SALICON for training models of goal-directed attention. Moreover, those suitably large datasets that do exist, each suffer from some weakness that limits their usefulness (\Tref{table: dataset cmp}), with the most common weakness being that the task used to collect the fixation behavior was not visual search (as in \cite{mathe2014actions, papadopoulos2014training}). Other datasets did use a search task, but either had people search for multiple targets simultaneously~\cite{gilani2015pet} or used only one target category (people; \cite{ehinger2009modelling}) or two (microwaves and clocks; \cite{zelinsky2019benchmarking}). All these  inadequacies demand  a new, larger, and higher-quality dataset of search fixations for model training. We use multiple fixation-based behavioral search metrics to interrogate COCO-Search18, which we attempt to predict using our IRL model and other state-of-the-art methods.

\myheading{Contributions}. This study makes several important contributions:
\textbf{(1)} We apply Inverse Reinforcement Learning (GAIL) to the problem of predicting fixation scanpaths during visual search, the first time this has been done for a goal-directed attention. 
\textbf{(2)} In order to apply IRL to scanpath prediction we needed to integrate changes in fixation location with changes in the state representation, a problem that we solved using Dynamic Contextual Beliefs. DCB is a novel state encoder that updates beliefs about peripherally-viewed objects (an object context) based on the movements of a simulated fovea. This is a technical novelty of our method. 
\textbf{(3)} We introduce COCO-Search18, a large-scale, high-quality dataset of COCO images annotated with the fixations of 10 people searching for 18 target-object categories. COCO-Search18 makes possible the deep network modeling of goal-directed attention. 
\textbf{(4)} We show through model comparison and with multiple metrics that our IRL model outperforms other state-of-the-art methods in predicting search scanpaths. We also show that the IRL model (i) learns an object's scene context; (ii) generalizes to predict the behavior of new subjects, and (iii) needs less data to achieve good performance compared to other models.
\textbf{(5)} Finally, we learned how to quantify a reward function for the fixations in a search task. This will make possible a new wave of experimental investigation that will ultimately result in a better understanding of goal-directed attention.

\section{Scanpath Prediction Framework}

\begin{figure}[t]
  \centering
  \includegraphics[width=.95\linewidth]{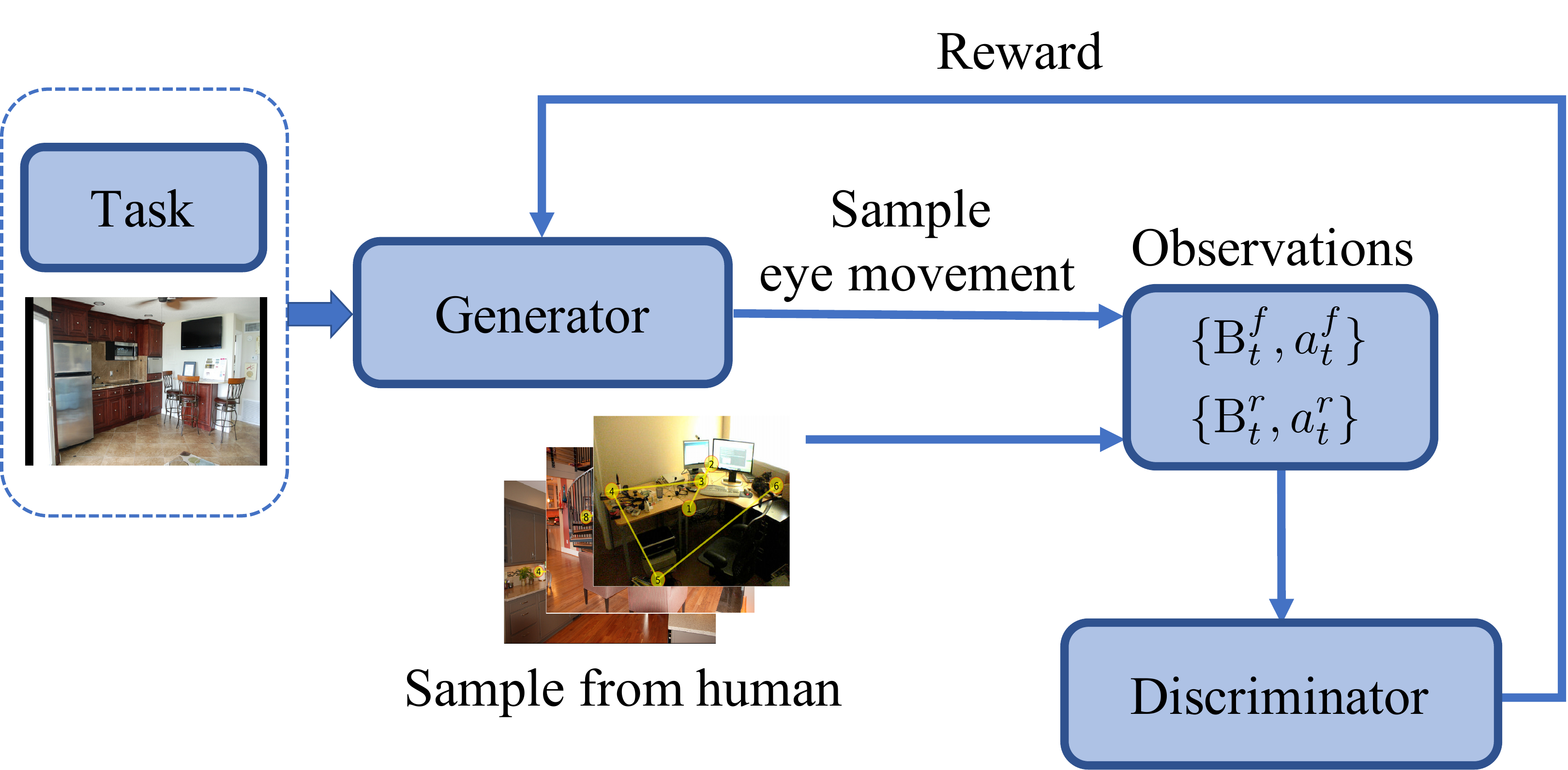}
  \vskip -0.1in
  \caption{{\bf Overview of the IRL framework}. The generator (policy) generates fake state-action pairs $\{B^f_t, a^f_t\}$ by sampling eye movements from given images and tasks. The discriminator (reward function) is trained to differentiates real human state-action pairs $\{B^r_t, a^r_t\}$ from the generated ones and provides reward to train the generator. The states $B^f_t$ and $B^r_t$ use DCB representations.}
  \label{fig:irl_pipeline}
  \vskip -0.15in
\end{figure}

We propose an IRL framework (Fig.~\ref{fig:irl_pipeline}) to model human visual search behavior. 
A person performing a visual search task can be considered a goal-directed agent, with their fixations being a sequential decision process of the agent. At each time step, the agent attends to (fixates) a specific location  within the image and receives a version of the image that is blurred to approximate the human viewer's visual state, what we call a retina-transformed image. This is an image that has high-resolution (non-blurred) information surrounding the attended location, and lower-resolution information outside of this central simulated fovea~\cite{perry2002gaze}. The state of the agent is determined by the sequence of visual information that accumulates over fixations toward the search target (\Sref{sec:state_rep}), with each action of the agent depending on the state at that time during the evolving state representation. The goal of the agent is to maximize internal rewards through changes in gaze fixation. While it is difficult to behaviorally measure how much reward is received from these fixations, with IRL this reward can be assumed to be a function of the state and the action, and this function can be jointly learned using the imitation policy (\Sref{sec:learning}). 

\begin{figure}[t]
  \centering
  \includegraphics[width=.95\linewidth]{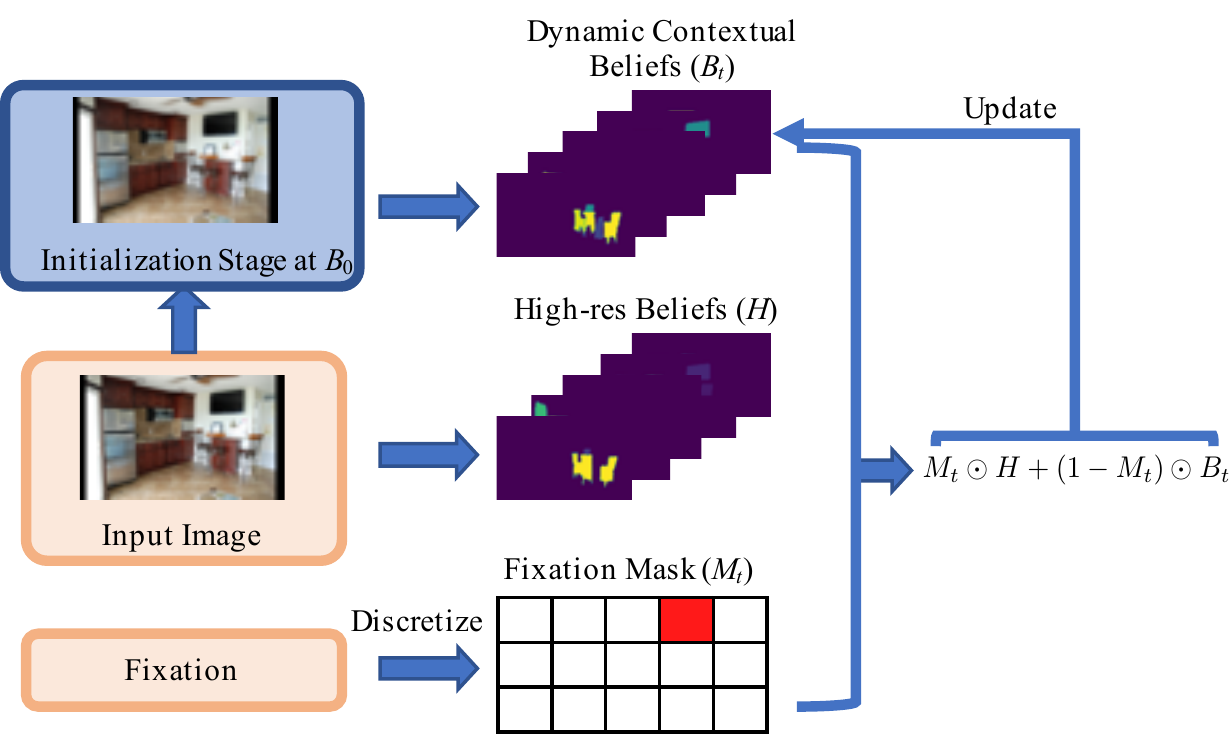}
  \vskip -0.1in
  \caption{{\bf Overview of DCB}. First, an input image and its low-res image counterpart are converted into the high-res beliefs and low-res beliefs. The initial state $B_0$ is set as the low-res belief. At each fixation which is discretized into a binary fixation mask $M_t$ with 1's at the fixation location and 0's elsewhere, a new state is generated by applying \Eref{eq:heuristic-update}.}
  \label{fig:belief_pipeline}
  \vskip -0.2in
\end{figure}


\subsection{State Representation Modeling}\label{sec:state_rep}


To model the state of a viewer we propose a novel state representation for accumulating information through fixations that we term a \textbf{Dynamic-Contextual-Belief (DCB)}. As shown in \Fref{fig:belief_pipeline}, DCB is composed of three components:~1) Fovea, which receives a high-resolution visual input only from the image region around the fixation location; 2) Contextual beliefs, which represent a person's gross ``what'' and ``where'' understanding of a scene in terms of level of class confidence; and 3) Dynamics, which actively collects information with each fixation made during search. We discuss each component in greater detail below.

\myheading{Fovea:}
The primate visual system has a fovea, which means that high-resolution visual information is available only at a central fixated location. To accumulate information from the visual world, it is therefore necessary to selectively fixate new image locations. Visual inputs outside of the fovea have lower resolution, with the degree of blur depending on the distance between the peripherally-viewed input and the fovea. Rather than implementing a full progressive blurring of an image input (i.e., a complete retina-transformed image, as in \cite{zelinsky2019benchmarking}), for computational efficiency here we use a local patch from the original image as the high-resolution foveal input and a blurred version of the entire image to approximate low-resolution input from peripheral vision.

\myheading{Contextual Belief:}
Attention is known to be guided to target (and target-like) objects during search, but more recently it has been suggested that attention is also guided to ``anchor objects'' \cite{boettcher2018anchoring, vo2019reading}, defined as those objects having a learned spatial relationship to a target that can help in the efficient localization of that target. For example, people often look at the wall when searching for a TV because TVs are often found hanging on the wall. Inspired by this, we propose to model, not only the target features (as in \cite{zhang2018finding}), but also other objects and background information in the state.

We hypothesize that people have an internal scene parser that takes an image input and generates belief maps for that image based on all the objects and background classes in that person's knowledge structure. We believe these belief maps also guide movements of the fovea for the purpose of capturing  high-resolution information and forming better beliefs. We approximate these belief maps using a Panoptic-FPN~\citep{kirillov2019panopticfpn} for panoptic segmentation \cite{kirillov2019panoptic}. Given an image, Panoptic-FPN generates a pixel-level mask for each ``thing'' class (object) and each ``stuff'' class (background) in the image. There are 80 ``thing'' categories (including a single ``other'' class for the 80 ``thing'' classes) and 54 ``stuff'' categories \cite{lin2014microsoft, caesar2018coco, kirillov2019panoptic}. We create a mask for each category by grouping all mask instances belonging to the same category and use the belief maps of the 134 categories as the primary component of the state representation. We term these belief maps {\it contextual beliefs} because the collective non-target beliefs constitute a context of spatial cues that might affect the selection of fixations during the search for a target. 


\myheading{Dynamics} refers to the change in the state representation that occurs following each fixation. 
We propose a simple yet effective heuristic to model state dynamics (see \Fref{fig:belief_pipeline}). Initially, the state is based on the contextual beliefs on the low-resolution image corresponding to a peripheral visual input. For each fixation by the searcher, we update the state by replacing the portion of the low-resolution belief maps with the corresponding high-resolution portion obtained at the new fixation location. The state is updated as follows:
\begin{align}
B_0 = L \ \textrm{and } B_{t+1} = M_t\odot H + (1-M_t)\odot B_t,    \label{eq:heuristic-update}
\end{align}
where $B_t$ is the belief state after $t$ fixations, $M_t$ is the circular mask generated from the $t^{th}$ fixation, $L$ and $H$ are the belief maps of ``thing'' and ``stuff'' locations for low-resolution and high-resolution images, respectively. Humans have different search behaviors on the same image given different search targets. To capture this, we augment the state by concatenating it with a one-hot task vector. Please refer to the supplementary material for more detail.



\subsection{Reward and Policy Learning}\label{sec:learning}
We learn the reward function and the policy for visual search behavior using Generative Adversarial Imitation Learning (GAIL) \cite{ho2016generative}. As shown in \Fref{fig:irl_pipeline}, GAIL is an adversarial framework with a discriminator and a generator. The policy is the generator that aims to generate state-action pairs that are similar to human behavior. The reward function maps a state-action pair to a numeric value, and this function is formulated as the logarithm of the discriminator output. The policy and reward functions are obtained by training the generator and discriminator with an adversarial optimization framework.

Let $D$ and $G$ denote the discriminator and the generator, respectively. The discriminator aims to differentiate human state-action pairs from fake state-action pairs generated by the policy. This corresponds to maximizing the following objective function: 
\begin{align}
\mL_{D} = & \mathbb{E}_r [\log(D(S, a))] + \mathbb{E}_f [\log(1-D(S,a))]  \nonumber \\
& - \lambda \mathbb{E}_r [\left\| \nabla D(S, a))\right\|^2]. 
\end{align}
In the above objective function, $\mathbb{E}_r$ denotes the expectation over the distribution of real state-action pairs, while $\mathbb{E}_f$ denotes the expectation over the fake samples generated by the generator (i.e., the policy). The last term of the above objective is the expected squared norm of the gradients, which is added for faster convergence \cite{roth2017stabilizing}. The reward function is defined based on the discriminator: 
\begin{equation}
	r(S, a) = \log (D(S, a)).
	\label{eq:rwd_fun}
\end{equation}

The generator aims to fool the discriminator, and its objective is to maximize the log likelihood of the generated state-action pairs, i.e., to maximize: $\mL_{G} = \mathbb{E}_f [\log(D (S, a))] = \mathbb{E}_f[r(S, a)]. $

%

The generator is an RL policy, hence its objective can be equivalently reformulated as an RL objective and optimized by Proximal Policy Optimization \cite{schulman2017proximal}:
\begin{align}
  &\mL_{\pi} = \mathbb{E}_{\pi} [\log(\pi(a|S)) A(S, a)] + H(\pi). 
\end{align}
We use GAE \cite{schulman2015high} to estimate the advantage function $A$ which measures the gain of taking action $a$ over the policy’s default behavior. $H({\pi}) = - \mathbb{E}_{\pi} [\log(\pi(a| S))]$, the entropy in max-entropy IRL \cite{Ziebart-etal-AAAI08}.

\section{COCO-Search18 Dataset}

COCO-Search18 is a large-scale and high-quality dataset of search fixations obtained by having 10 people viewing~6202 images in the search for each of 18 target-object categories. Half of these images depicted an instance of the designated target object and the other half did not, meaning that we adopted a standard target-present (TP) or target-absent (TA) search task. All images in COCO-Search18 were selected from the COCO trainval set \cite{lin2014microsoft}. Five criteria were imposed when selecting the TP images: 
\textbf{(1)} No images depicting a person or an animal (to avoid known strong biases to these categories that might skew our measures of attention control~\cite{cerf2008predicting, judd2009learning}). 
\textbf{(2)} The image should include one and only one instance of the target.
\textbf{(3)} The size of the target, measured by the area of its bounding box, must be $>$1\% and \textless10\% of the area of the search image. 
\textbf{(4)} The target should not be at the center of the image, enforced by excluding an image if the target bounding box overlapped with the center cell of a 5x5 grid. 
\textbf{(5)} The original image ratio (width/height) must be between 1.2 and 2.0 to accommodate the display screen ratio of 1.6. After applying these exclusion criteria, and excluding object categories that had less than 100 images of exemplars, we were left with 32 object categories (out of COCO's 80) to use as search targets. To exclude images in which the target was highly occluded or otherwise difficult to recognize, we trained a patch-based classifier for target recognition and only selected images in which the cropped target-object patch had a classification confidence in the top 1\%. Finally, we manually excluded images depicting digital clocks from the clock target category (because the features of analog and digital clocks are very different and this would be expected to reduce data quality by creating variability in the search behavior), as well as images depicting objectionable content. This left us with 3101 TP images over 18 target categories. To select the same number of TA images for each of these 18 categories, we randomly sampled COCO trainval images with the following constraints: 
\textbf{(1)} The image should not depict an instance of the target, and \textbf{(2)} The image must include at least two instances of the target's siblings, as defined in COCO. For example, a microwave sibling can be an oven, a toaster, a refrigerator, or a sink, which are under the parent category of appliance. We did this to discourage TA responses from being based on scene type (e.g., a city street scene would be unlikely to contain a microwave).

\begin{figure}
\centering
\includegraphics[width=1.\linewidth]{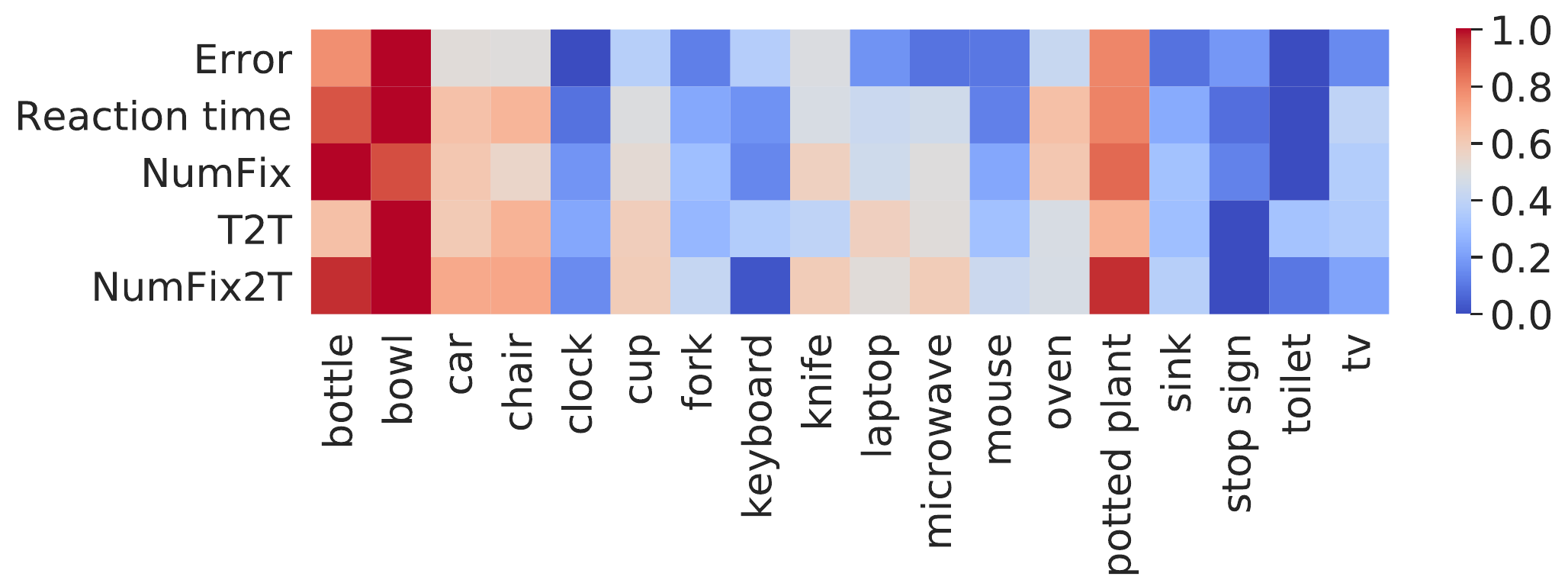}
\vskip -0.1in
\caption{Normalized gaze data [0,1] on response error, reaction time, number of fixations (NumFix), time to target (T2T), and number of fixations to target (NumFix2T) averaged over 10 subjects searching for 18 categories in TP images. Redder color indicates harder search targets, bluer color indicates easier search.}
\label{fig:gaze_data}
\vskip -0.2in
\end{figure}

Each of the 10 student participants (6 males, age range 18-30, normal or corrected-to-normal vision) viewed all 6202 images, and their eye position throughout was sampled at 1000Hz using an EyeLink 1000 eyetracker (SR Research) in tower-mount configuration under controlled laboratory conditions. For each subject, data collection was distributed over six sessions in six days, with each session having equal number TP trials and TA trials ($\sim$500 each) randomly interleaved. Each session required $\sim$2 hours. For each image, subjects made a TP or TA judgment by pressing a `yes' or `no' button on a game pad. They searched all the images for one target category before preceding to next category. A total of 299,037 fixations were extracted from the eye position data, over the 10 subjects, although only data from the TP fixations will be reported here (\Fref{fig:gaze_data}). TP fixations occurring on error trials, or after fixation on the target, were discarded. This left 100,232 TP search fixations to use for training and testing. 
All model evaluations are based on 70\% training, 10\% validation, and 20\% test, random splits of COCO-Search18, within each target category. 

\section{Experiments}
We evaluate the proposed framework and its constituent components in multiple experiments.  We first compare the scanpath predictions by the IRL-based algorithm to predictions from various heuristic methods and  behavior cloning methods using ConvNets and convolutional LSTM. We then study the algorithm's ability to generalize to new human subjects. Finally, we analyze a context  effect, the value of having more training data, and report the results of an ablation study. Here we used only the target-present trials from COCO-Search18, leaving analyses of the target-absent data for future study.



\subsection{Comparing Scanpath Prediction Models}\label{sec:policy_cmpr}
\myheading{Comparison methods}. We compare the IRL algorithm for predicting scanpaths to several baselines, heuristics, and behavior cloning methods: (1) \textbf{Random scanpath}: we predict the scanpath for an input image by randomly selecting a human scanpath for the same search target but in a different input image. (2) \textbf{Detector}: we sample a sequence of fixation locations based on the detection confidence map over the image. Regions with higher confidence scores are more likely to be sampled. (3) \textbf{Fixation heuristics}: rather than sampling from a detector's confidence map, here we generate fixations by sampling from a  fixation density map produced by a ConvNet trained on human fixation density maps.
(4) {\bf BC-CNN} is a behavior cloning method, where we train a ConvNet to predict the next fixation location from the DCB state representation. Note that this state representation and network structure are identical to the one used by the IRL policy described in \Sref{sec:state_rep}. (5) {\bf BC-LSTM} is a behavior cloning method similar to BC-CNN, but the state representation and update are done with a convolutional LSTM. Instead of having the simple predefined update rule used by both IRL and BC-CNN, as shown in \Eref{eq:heuristic-update}, BC-LSTM aims to learn a recurrent update rule automatically with an LSTM: $B_{t+1} = ConvLSTM(B_t, I_t)$, where $ConvLSTM$ denotes a convolutional LSTM cell~\cite{ballas2015delving}, $B_t$ is the hidden state of the LSTM cell and also the searcher's belief state after $t$ fixations.  $I_t$ is the input to the LSTM at time $t$, and it is defined as $I_t = M_t\odot H + (1-M_t)\odot L$. Recall that $M_t$ is the circular mask generated from the $t^{th}$ fixation, $L$ and $H$ are the predicted maps from the Panoptic-FPN \cite{kirillov2019panoptic} for the 80 COCO objects and 54 ``stuff'' classes for low- and high-resolution input images, respectively. 








\begin{figure}[t]
  \centering
  \includegraphics[width=.9\linewidth]{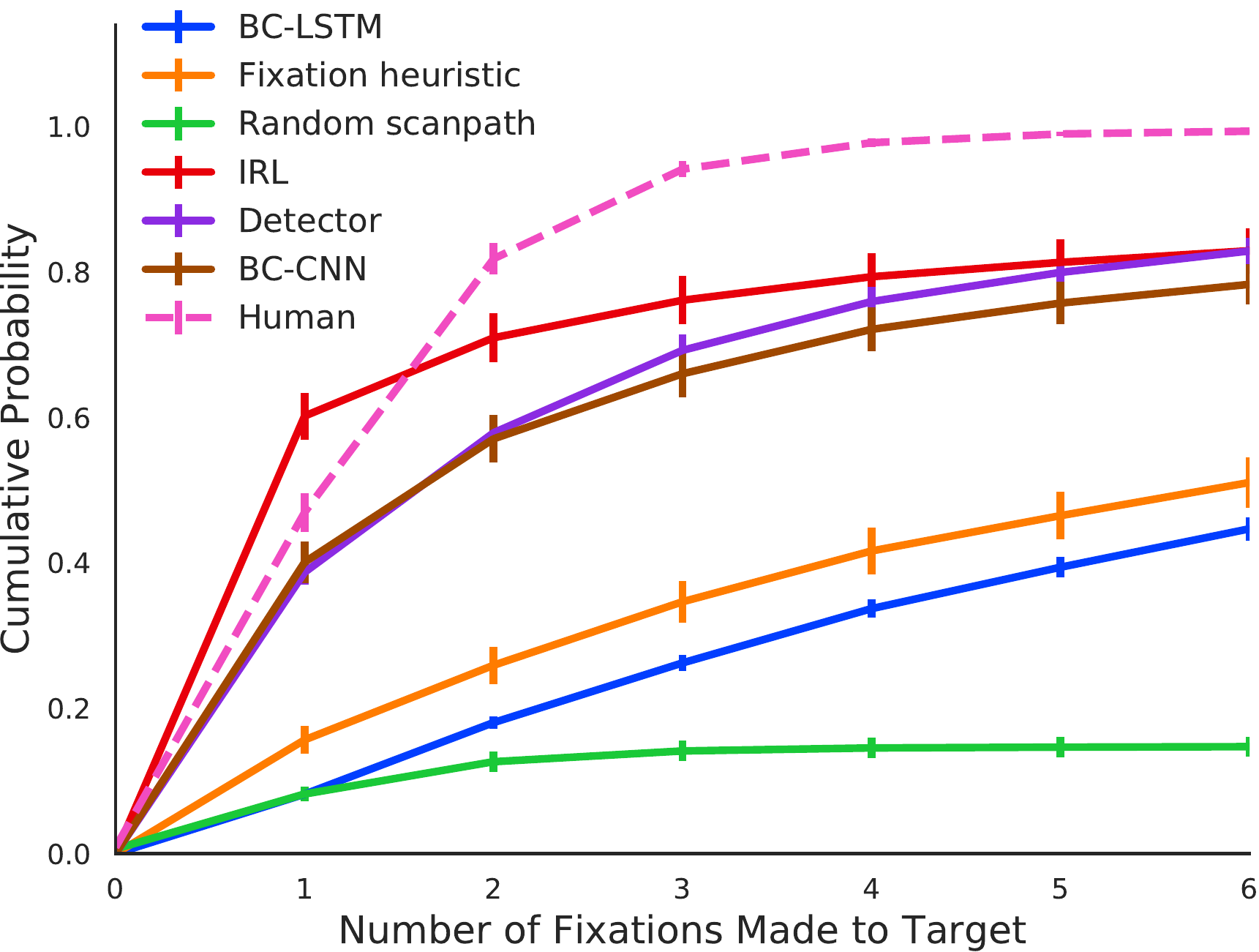}
  \vskip -0.1in
  \caption{{\bf Cumulative probability of fixating the
target} for human searchers and all predictive methods. X-axis is the number of fixations until the fovea moves to the target object; Y-axis is the percentage of scanpaths that succeed in locating the target. Means and standard errors are first computed over target categories, and then over searchers.}
  \label{fig:cdf}
\vskip -0.2in
\end{figure}
\myheading{Results}. \label{sec:results}
Fig. \ref{fig:cdf} shows the cumulative probability of gaze landing on the target after each of the first 6 fixations made by humans and the algorithms in our model comparison. First, note that even the most predictive models have a performance ceiling lower than that of humans, whose ceiling over this range is nearly 1. These lower ceilings likely reflect a proportion of trials in which the models search was largely unguided. Second, note the steep increase in  target fixation probability after the first and second fixations. The slopes of these functions indicate strong target guidance. The target was fixated in the very first movement on about half of the images, with the IRL model replicating human search guidance slightly better than its nearest competitors: the Detector and BC-CNN models. 

We quantify the patterns from Fig. \ref{fig:cdf} using several metrics. Two of these metrics follow directly from Fig. \ref{fig:cdf} and capture aggregate measures combining search guidance and accuracy. The first of these computes the area under the cumulative probability of the target fixation curve, a metric we refer to as {\bf Target Fixation Probability AUC} or TFP-AUC. Second, we compute the sum of the absolute differences between the human and  model cumulative probability of target fixation in a metric that we refer to as {\bf Probability Mismatch}. We also report the {\bf Scanpath Ratio}, which is a widely used metric for search efficiency. It is computed by the ratio of Euclidean distance between the initial fixation location and the center of the target to the summed Euclidean distances between fixations to the target~\citep{hout2015target}. Finally, we compute two metrics for scanpath prediction success, that both  capture the scanpath similarity between fixation sequences generated by humans and  sequences generated by the model. The first of these computes a {\bf Sequence Score} by first converting a scanpath into a string of fixation cluster IDs and then use a string matching algorithm \cite{Needleman-Wunsch-JMB70} to measure similarity between two strings. Finally, we use {\bf MultiMatch}~\cite{anderson2015comparison, dewhurst2012depends} to measure the scanpath similarity at the pixel level. MultiMatch measures five aspects of scanpath similarity: shape, direction, length, position, and duration. We exclude the duration metric because the studied models do not predict fixation duration. Unless otherwise specified, each model generates 10 scanpaths of maximum length 6 (excluding the first fixation) for each testing image by sampling from the predicted action map at each fixation, with the results averaged over scanpaths.

\begin{figure}[t]
\begin{center}
\includegraphics[width=.48\textwidth]{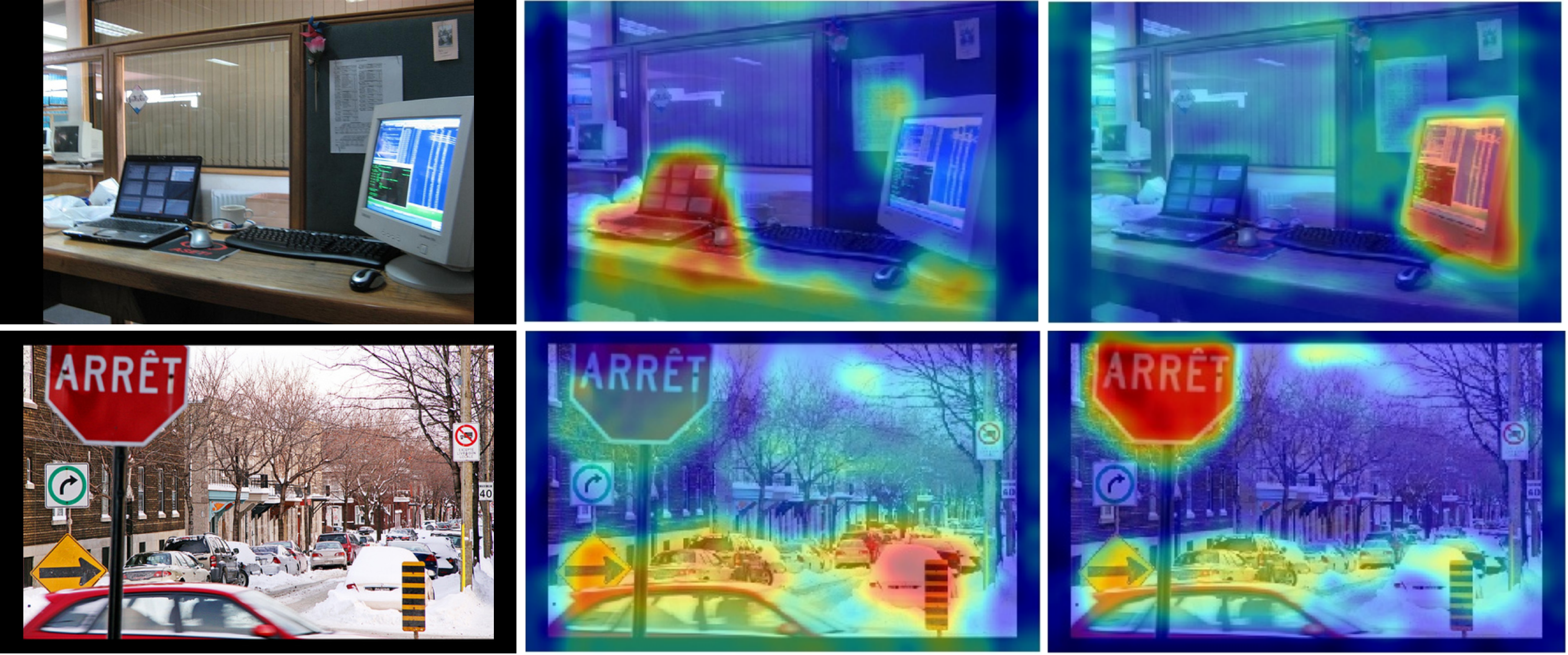}
\end{center}
\vskip -0.2in
  \caption{Initial {\bf reward maps} learned by the IRL model for two different search targets in two test images. Top row: original image (left), mouse target (middle), and tv target (right). Bottom row: original image (left), car target (middle), stop sign target (right). Redder color indicates the expectation of higher reward for fixating a location.
}
\label{fig:rwd_maps}
\vskip -.2in
\end{figure}

As  seen from \Tref{table:all_results}, the IRL algorithm outperforms the other methods on all metrics. The performance of IRL is closest to {\it Human}, an oracle method where the scanpath of a subject is used to predict the scanpath of another subject for the same input image. \Fref{fig:rwd_maps} also shows that {\bf reward maps} recovered by the IRL model depend greatly on the category of the search target. In the top row, higher reward was assigned to the laptop when searching for a mouse, while for the same image greater reward was expected from fixating on the monitor when searching for a tv. Similarly, the search for a car target in the bottom image resulted in the expectation of reward from the other cars on the road but almost not at all from the highly-salient stop sign, which becomes intensely prioritized when the stop sign is the target. 

\begin{table*}[t]
\begin{center}

\begin{tabular}{l|c|c|c|c|cccc}
 & \multirow{2}{*}{TFP-AUC $\bm{\uparrow}$} & \multirow{2}{1.8cm}{Probability Mismatch $\bm{\downarrow}$} & \multirow{2}{1.5cm}{Scanpath Ratio $\bm{\uparrow}$} & \multirow{2}{1.5cm}{Sequence Score $\bm{\uparrow}$} & \multicolumn{4}{c}{MultiMatch $\bm{\uparrow}$} \\ \cline{6-9} 
 &  &  &  &  & shape & direction & length & position \\ \Xhline{1.2pt}
Human & 5.200 & - & 0.862 & 0.490 & 0.903 & 0.736 & 0.880 & 0.910 \\ \hline
Random scanpath & 0.795 & 4.407 & - & 0.295 & 0.869 & 0.558 & 0.849 & 0.849 \\
Detector & 4.046 & 1.166 & 0.687 & 0.414 & 0.877 & 0.676 & 0.853 & 0.863 \\
Fixation heuristic & 2.154 & 3.046 & 0.545 & 0.342 & 0.873 & 0.614 & \textbf{0.870} & 0.850 \\
BC-CNN & 3.893 & 1.328 & 0.706 & 0.409 & 0.880 & 0.669 & 0.865 & 0.874 \\
BC-LSTM & 1.702 & 3.497 & 0.406 & 0.324 & 0.834 & 0.567 & 0.818 & 0.770 \\
IRL(Ours) & \textbf{4.509} & \textbf{0.987} & \textbf{0.826} & \textbf{0.422} & \textbf{0.886} & \textbf{0.695} & 0.866 & \textbf{0.885}
\end{tabular}

\end{center}
\vskip -0.2in
\caption{ {\bf Comparing scanpath prediction algorithms} (rows) using multiple scanpath metrics (columns) on the COCO-Search18 test dataset. In the case of Sequence Score and Multimatch, \enquote{Human} refers to an oracle method where one searcher's scanpath is used to predict another searcher's scanpath; \enquote{Human} for all other metrics refers to observed behavior.}
\label{table:all_results}
\vskip -.2in
\end{table*}



\myheading{Implementation details.}
We resize each input image  to $\htimesw{320}{512}$ and obtain a low-resolution image by applying a Gaussian filter with standard deviation \mbox{$\sigma = 2$}. To compute the contextual beliefs, we use a Panoptic-FPN with backbone network ResNet-50-FPN pretrained on COCO2017 \cite{kirillov2019panoptic}. Panoptic-FPN outputs a feature map of 134 channels, corresponding to 80 object categories and 54 background classes in COCO, and it is resized to $\htimesw{20}{32}$ spatially. 

For IRL and BC-CNN, we use the same policy network architecture: a network composed of four convolutional (conv) layers and a softmax layer.
IRL model has two additional components---critic network and discriminator network. The \textbf{critic network} has two convolutional layers and two fully-connected (fc) layers. The discriminator network shares the same sturcture with the IRL policy network except the last layer which is a sigmoid layer.
Each conv layer and fc layer in BC-CNN and IRL is followed by a ReLU layer. 
BC-LSTM has the same policy network as the BC-CNN, with the difference being the use of a convolutional LSTM~\cite{ballas2015delving} to update the states. BC-CNN and BC-LSTM use the KL divergence between predicted spatial distribution and ground truth as loss.
The prediction of both behavior cloning models and IRL is conditioned on the search target. We implement the target conditioning by introducing an additional bias term based on the search task to the input features at each layer \cite{perez2018film}. The human visual system employs Inhibition-of-Return~(IOR) to spatially tag previously attended locations with inhibition to discourage attention from returning to a region where information has already been depleted~\cite{wang2010searching}. To capture this mechanism, we enforce IOR on the policy by setting the predicted probability map to 0 at each attended location using a \htimesw{3}{3} grid.
See the supplementary for more detail.

\subsection{Group Model vs Individual Model}
\begin{figure}[t!]
\begin{center}
\includegraphics[width=.48\textwidth]{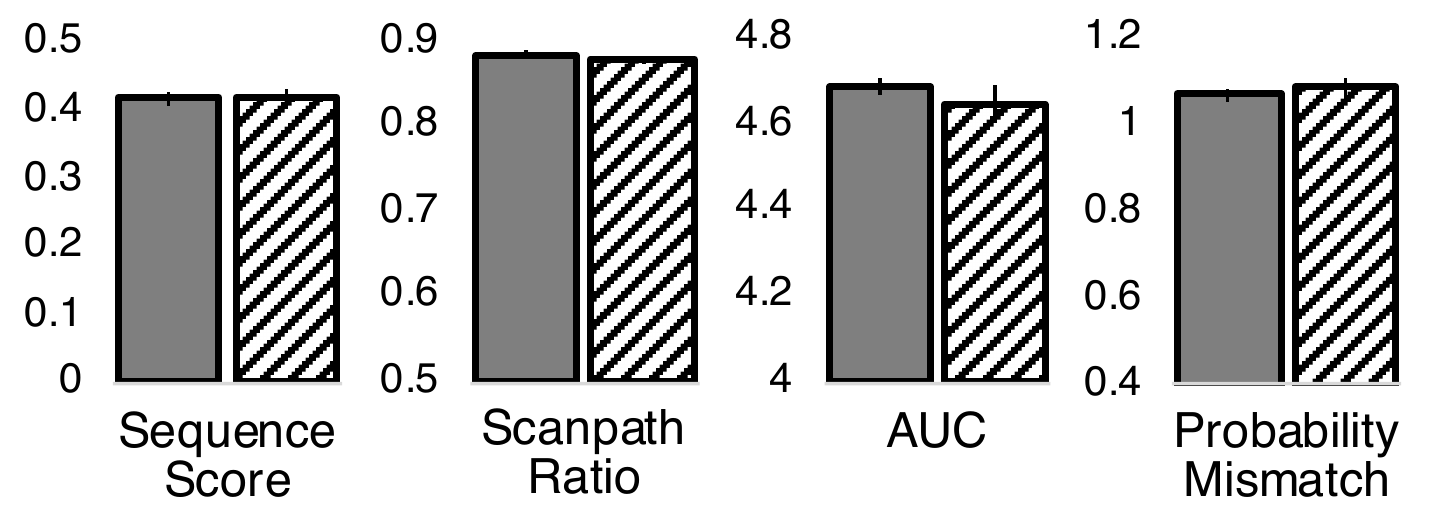}
\end{center}
\vskip -0.25in
  \caption{No significant differences were found between a group model (solid), trained with 9 subjects, and an individual model (striped), trained with one subject. }
\label{fig:group vs indv}
\vskip -0.2in
\end{figure}

The previous subsection described the IRL model's ability to predict a searcher's scanpath on unseen test images, but how well can this model predict the scanpaths of a new unseen searcher without training on that person's scanpaths? To answer this question, we perform ten leave-one-subject-out experiments, with each experiment corresponding to a test subject. For every subject we train two models: (1) a group model using the scanpaths of the 9 other subjects; and (2) an individual model using the scanpaths of the test subject on the training images. We evaluate the performance of these models on the scanpaths of each test subject on the unseen test images. \Fref{fig:group vs indv} shows that both models  perform well, with an insignificant performance gap between them. This suggests that there is good agreement between group and individual behaviors, and that a group model can generalize well to new searchers.

\subsection{Context Effects}\label{sec:context}
\begin{figure}
\centering
\includegraphics[width=1.0\linewidth]{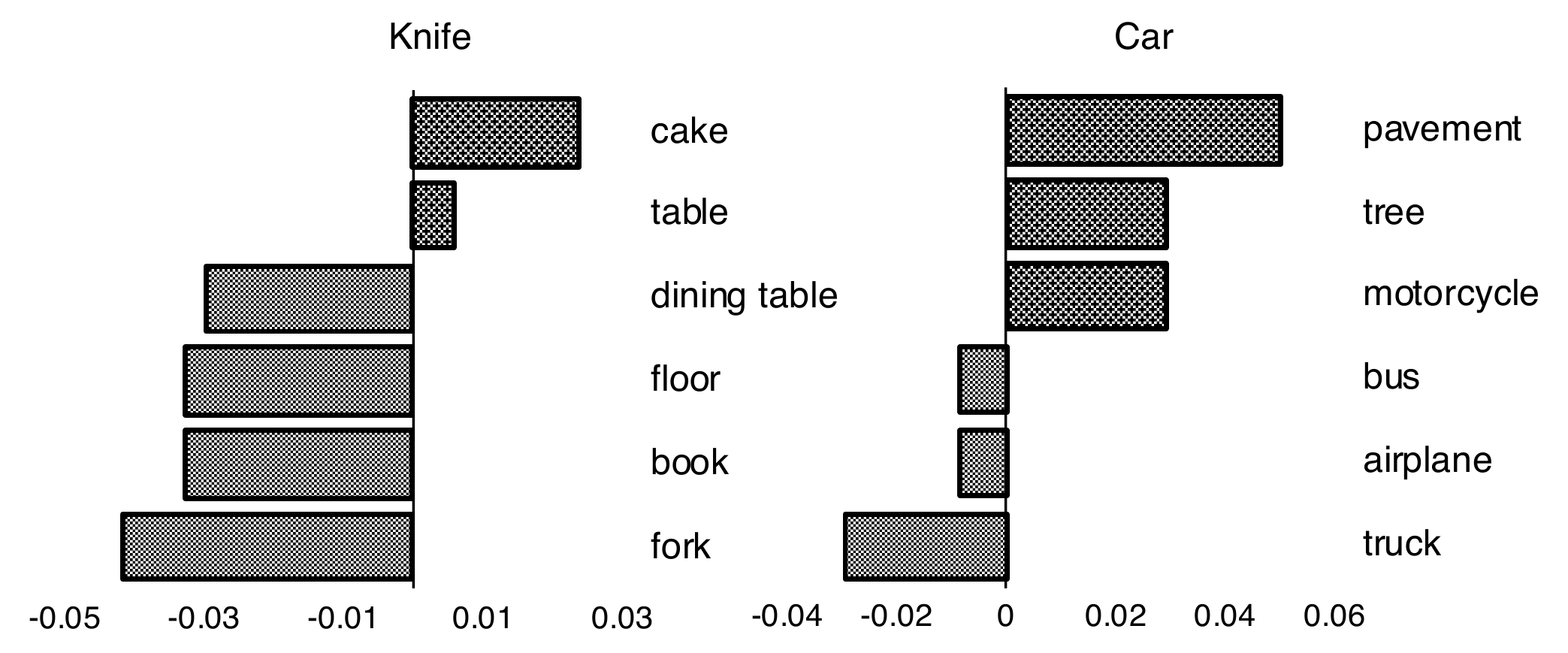}
  \vskip -0.1in
  \caption{{\bf Context effect}. The six most influential context objects (grey bars) for knife and car search tasks. The y-axis is the context object category and the x-axis is a measure of how much the belief map for a context object contributed to search efficiency, as measured by TFP-AUC. Larger positive values mean that the context object improved search guidance to the target, more negative values mean that the object distracted attention from the search.}
  \label{fig:ctx_effect}
  \vskip -0.1in
\end{figure}

\myheading{Search efficiency}. With DCB we can ask how an object from category $A$ affects the search for a target from category~$B$. This effect can either increase (guidance) or decrease (distraction) search efficiency. To study this, we first zero out the belief map of category $A$ in the DCB state representation and then measure the TFP-AUC (see \Sref{sec:results}) on test images for category $B$. We compute the difference between the TFP-AUC obtained with and without switching off the belief map for category $A$ in DCB.  A positive value indicates that an object in category $A$ helps to guide search for a target in category $B$, while a negative value indicates the opposite (that the object is distracting). We did this for the 134 COCO objects and stuff non-target categories $A$ and the 18 target categories $B$.  \Fref{fig:ctx_effect} shows the six most guiding and  distracting objects for the knife and car searches. Note that the fork was highly distracting when searching for a knife, likely because the two look similar in periphery vision, but that the cake facilitated the knife search. Similarly for the car search, pavement provided the strongest guidance whereas trucks were the  most distracting.

\begin{figure}[t]
  \centering
  \includegraphics[width=1.0\linewidth]{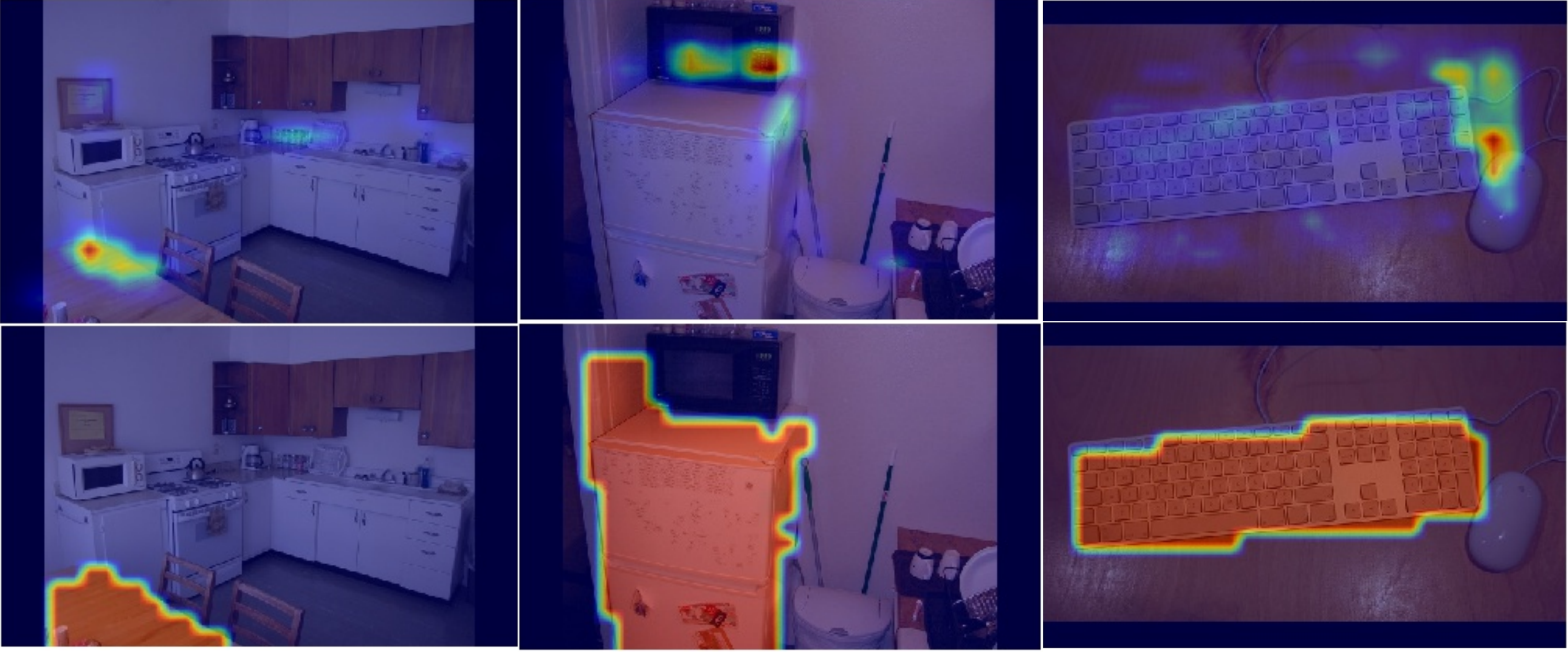}
  \vskip -0.1in
  \caption{{\bf Spatial relations} between context and target objects learned by the model. Top row shows individual context maps for a dining table (left) and a refrigerator (middle) in a microwave search, and a keyboard (right) in a mouse search. Bottom row are the belief maps of the corresponding context objects. Gaze is guided to the top of the dinning table and refrigerator when searching for a microwave, and to the right of the keyboard when searching for a mouse.}
   \vskip -0.15in
  \label{fig:spatial}
\end{figure}

\myheading{Directional Prior}. Can an object from category $A$ serve as a directional spatial cue in the search for a target from category $B$? Suppose $M$ is the probability map produced by the policy network of our IRL model, and let $M'$ be the modified probability map from the policy network but with the belief map of category $A$ in the DCB state representation being switched off. By computing the difference between~$M'$ and~$M$ which we call a \textit{context map} (as depicted in the top row of \Fref{fig:spatial}), we can see the spatial relationship between the context object $A$ and the target object $B$ (see \Fref{fig:spatial} for examples).


\subsection{Ablation Study on State Representation}\label{sec:state_cmpr}

DCB is a rich representation that uses top-down, bottom-up, and history information. Specifically, it consists of 136 belief maps, divided into five factor groups: target object (1 map), context objects (79 maps), ``stuff'' (54 maps), saliency (1 map, extracted using DeepGaze2 \cite{kummerer2017understanding}), and history (1 binary map for the locations of previous fixations). To understand the contribution of a factor group, we remove the group from the full state representation and measure the effect on performance. As we describe in the supplementary material, the most important factor groups were target and context objects, followed by stuff, whereas saliency and history weakly impacted model performance. In addition,
an alternative state representation to DCB is the Cumulative Foveated Image (CFI)~\cite{zelinsky2019benchmarking}, but replacing DCB with CFI degrades the performance of IRL.
Full results can be found in the supplementary material. 

\subsection{Data Efficiency}

\begin{figure}
\centering
\includegraphics[width=1.0\linewidth]{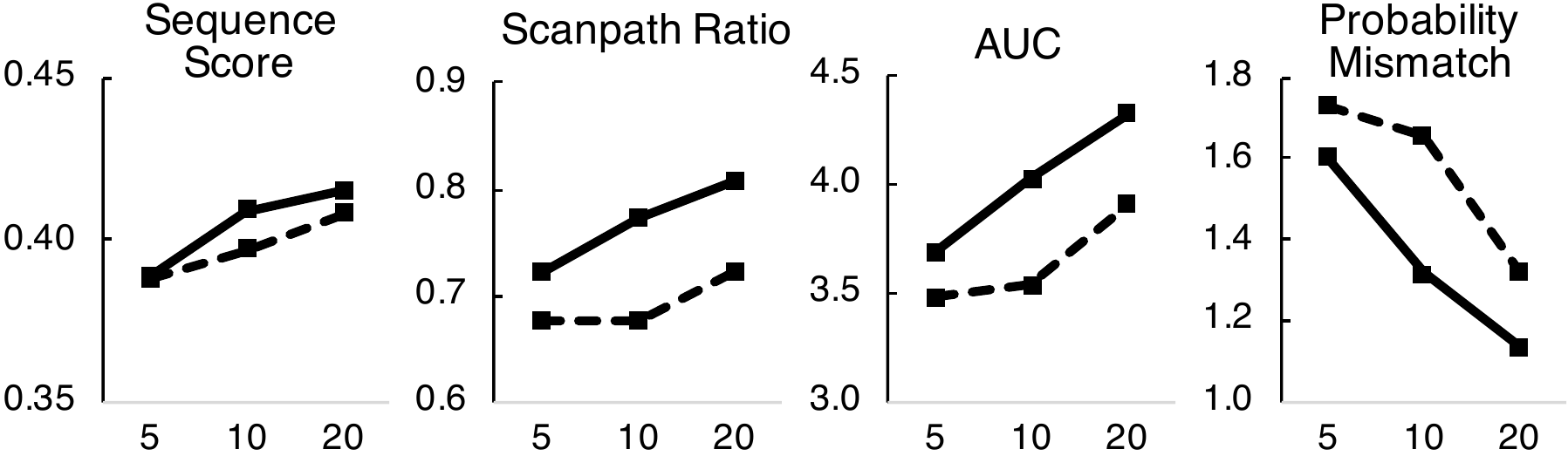}
\vskip -0.1in
  \caption{Performance of IRL (solid line) and BC-CNN (dashed line) as the number of training images per category increases from 5 to 20. IRL is more data efficient than BC-CNN, likely due to an adversarial data generator.}
  \label{fig: data_eff}
  \vskip -0.2in
\end{figure}

\Fref{fig: data_eff} shows  IRL and BC-CNN performance as we vary the number of training images per object category. Both methods use DCB as the state representation. IRL is more data efficient than BC-CNN, achieving  comparable or better results using less training data. A likely reason for this is that the GAIL-based~\cite{ho2016generative} IRL method includes an adversarial component that generates augmented training data, leading to a   less prone to overfitting policy network. Data efficiency is crucial for training  for new categories, given the time and cost of collecting human fixations. 

\section{Conclusions}

We proposed a new model for predicting search fixation scanpaths that uses IRL to jointly recover the reward function and policy used by people during visual search. The IRL model uses a novel and highly explainable state representation, \textit{dynamic contextual beliefs (DCB)}, which updates beliefs about objects to obtain an object context that changes dynamically with each new fixation. To train and test this model we also introduced COCO-Search18, a large-scale dataset of images annotated with the fixations of people searching for target-object goals. Using COCO-Search18, we 
showed that the IRL model outperformed comparable models in predicting search scanpaths. 

Better predicting human search behavior means better robotic search applications and human-computer systems that can interact with users at the level of their attention movements \cite{park2019advancing}. It may also be possible to use reward maps from the IRL model to annotate and index visual content based on what is likely to attract a person's attention. Finally, our work impacts the behavioral vision literature, where the visual features guiding human goal-directed attention are still poorly understood for real images \cite{zelinsky2013modeling}.

\myheading{Acknowledgements}. This project is supported by US National Science Foundation Award IIS-1763981, the Partner University Fund, the SUNY2020 Infrastructure Transportation
Security Center, and a gift from Adobe.

{\small
\setlength{\bibsep}{0pt}
\bibliographystyle{plainnat}

\bibliography{longstrings,all_pubs,pubs}
}

\cleardoublepage

\appendix
\section{Appendix}
This section provides further details about the COCO-Search18 dataset (\Sref{sec:dataset}), Dynamic Contextual Beliefs (\Sref{sec:DCB}), and implementation (\Sref{sec:impl_details}). We also include additional results from experiments and ablation studies, and interpretation (\Sref{sec:results}).

\subsection{Details about COCO-Search18 Dataset}\label{sec:dataset}
\myheading{Data source:}
The COCO-Search18 dataset annotates COCO \cite{lin2014microsoft} with human gaze fixations made during a standard target-present (TP) or target-absent (TA) search task, where on each trial the search image either depicted the target (TP) or it did not (TA). All of the images were selected from the {\it trainval} set, and detailed descriptions of TP and TA image selection and gaze collection methods are provided below.

\myheading{Target present image selection:} 


In addition to the exclusion criteria described in the main text, we also excluded images in which the target was highly occluded or otherwise difficult to recognize. Specifically, we only selected images in which the cropped target-object patch had a classification confidence $>$.99. 
To train this classifier, we cropped the target in each image (by bounding box) and used these image patches as positive samples. Same-sized image patches of non-target objects were used as negative samples. Negative samples were constrained to intersect with the target by 25\% (area of intersection divided by area of target) so that they could serve as hard negatives for specific targets. More than 1 million cropped patched were collected and resized to 224x224 pixels, while keeping the original aspect ratio by padding. The classifier is fine-tuned from an ImageNet-pretrained ResNet-50 model with the last fully connected layer changed from 1000 outputs to 33 (32+``Negative''). Images with a classification score for the cropped target patch that was \textless.99 were excluded. This resulted in 18 categories with at least 100 images in each category, and 3131 images in total. As described in the main text, we conducted a final manual checking of the dataset to exclude images depicting digital clocks (5 images), so as to make the clock target category specific to analog clocks, and to remove images depicting content that participants might find objectionable. This latter criterion resulting in the exclusion of 30 images, 22 of which were from the toilet category. 


\begin{table}
\begin{center}
\begin{tabular}{lcccc}
\toprule 
Category & TP images & ACC & TA images & ACC \\
\midrule 
bottle & 166 & 0.84 & 166  & 0.92 \\
bowl & 141 &0.80 & 141  &  0.90\\
car & 104 &0.89 & 104  & 0.91\\
chair & 253 &0.89 & 253  & 0.64\\
clock & 119 &0.99 & 119  & 0.97\\
cup & 276 &0.92 & 276  & 0.76\\
fork & 230 &0.96 & 230  & 0.98\\
keyboard & 184 &0.92 & 184  & 0.98\\
knife & 141 &0.89 & 141  & 0.97\\
laptop & 123 &0.95 & 123  &  0.95\\
microwave & 156 &0.97 & 156  & 0.95\\
mouse & 109 &0.97 & 109  & 0.97\\
oven & 101 &0.91 & 101  & 0.93\\
potted plant & 154 &0.84 & 154  & 0.95\\
sink & 279 &0.97 & 279  & 0.94\\
stop sigh & 126 &0.95 & 126  & 0.99\\
toilet & 158 &0.99 & 158  & 1.00\\
tv & 281 &0.96 & 281  & 0.93\\
\midrule
total/mean & 3101 & 0.92 & 3101 & 0.92\\
\bottomrule 
\end{tabular}
\caption{Number of images and response accuracy (ACC) for TP and TA images grouped by target category.}
\label{table: image numbers}
\end{center}
\end{table}

\begin{figure*}[t]
\centering
\includegraphics[width=1.\linewidth]{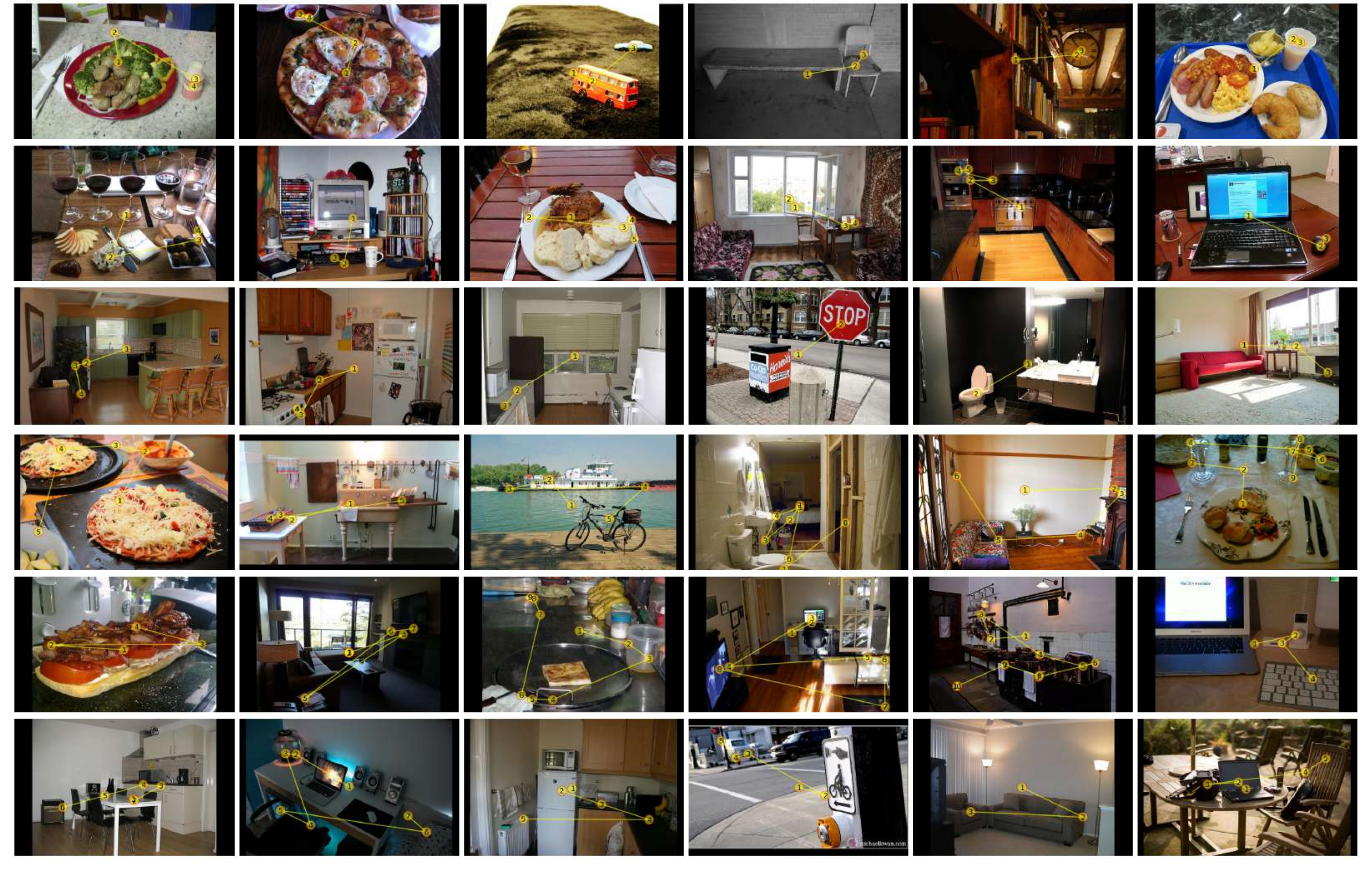}
\vskip -0.05in
\caption{Examples of human scanpaths during target-present (top 3 rows) and target-absent (bottom 3 rows) visual search. From left to right and top to bottom, the 18 target categories are: bottle, bowl, car, chair, clock, cup, fork, keyboard, knife, laptop, microwave, mouse, oven, potted plant, sink, stop sign, toilet, and tv. Each yellow line represents the scanpath of one behavioral searcher, with numbers indicating fixation order. }
\label{fig:scanpath_example}
\end{figure*}

After implemented all exclusion criteria, we selected 3101 target-present images from 18 categories: bottle, bowl, car, chair, clock, cup, fork, keyboard, knife, laptop, microwave, mouse, oven, potted plant, sink, stop sign, toilet, tv. See Table \ref{table: image numbers} for the specific number of images in each category and the average response accuracy (ACC). There were an equal number of TA images (for a total of 6202 images), which were all resized and padded to fit the ${1050}\times{1680}$ resolution of the display monitor.


\myheading{Gaze data collection procedure:}
Ten university undergraduate and graduate students (6 males, age range 18--30) with normal or corrected to normal vision participated in this study, which was approved by the Institutional Review Board. They were naive with respect to experimental question and design, and were compensated with course credits or money for their participation. Informed consent was obtained at the beginning of the experiment, and every participant read and understood the consent form before signing it.

The 6202 images were divided into six days of experiment sessions with each session consisting of $\sim$500 TP images and the same number of TA images, randomly interleaved. Images for a given target category were grouped and presented sequentially in an experiment block (i.e., target type was blocked). Preceding each block was a calibration procedure needed to map eye position obtained from the eye-tracker to screen coordinates, and a calibration was not accepted until the average calibration error was $\leq$.51 and the maximal error was $\leq$.94. Each trial began with a fixation dot appearing at the center of the screen. To start a trial, the subject should press the ``X'' button on a gamepad while carefully looking at the fixation dot. A scene would then be displayed and their task was to answer ``yes'' or ``no'' whether an exemplar of the target category for that block appears in the displayed scene. The subject registered a ``yes'' target judgment by pressing the right rigger of the gamepad, and a ``no'' judgment by pressing the left trigger. They were told that there were equal number of target present and absent trials, and that they should respond as quickly as possible while remaining accurate. Participants were allowed to take multiple breaks between and within each block.

Image presentation and data collection was controlled by Experiment Builder (SR research Ltd., Ottawa, Ontario, Canada). Images were presented on a 22-inch LCD monitor (resolution: 1050x1680), and subjects viewed these stimuli in a distance of 47cm from the monitor, enforced  by both chin rest and head rest. Eye movements were recorded using an EyeLink 1000 eye tracker in tower-mount configuration (SR research Ltd., Ottawa, Ontario, Canada). The experiment was conducted in a quiet and dimmed laboratory room. \Fref{fig:scanpath_example} shows some TP and TA images from the 18 object categories, with overlaid human scanpaths.

\subsection{Detailed Description of DCB}\label{sec:DCB}

\begin{figure*}[t]
  \centering
  \includegraphics[width=1.0\linewidth]{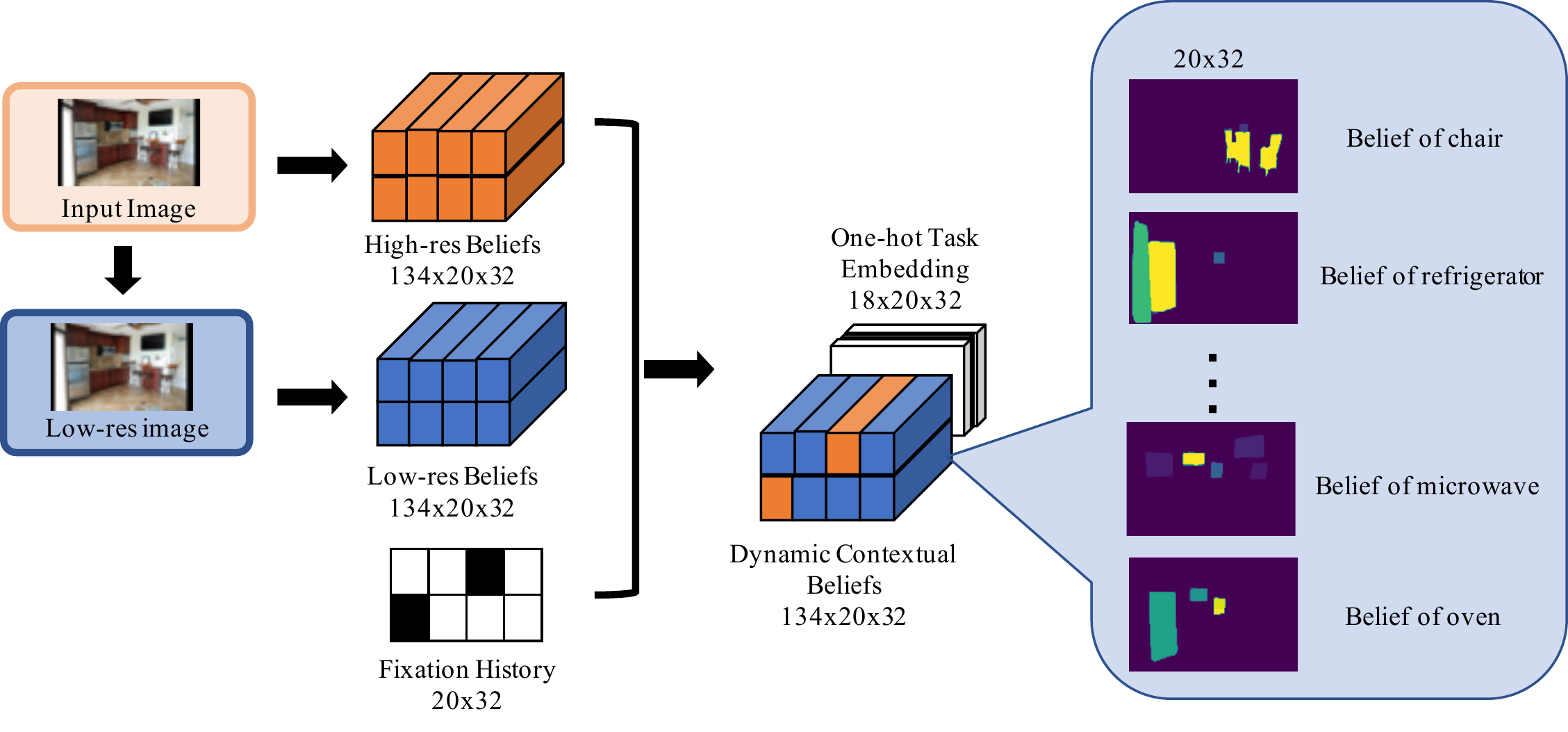}
  \vskip -0.1in
  \caption{{\bf Detailed illustration of Dynamic-Contextual-Belief}. First, an input image and its low-res image counterpart are converted into high-res beliefs and low-res beliefs. At each fixation, which is discretized into a binary fixation history map with 1's around the fixation location and 0's elsewhere, a new state is generated by concatenating the output of \Eref{eq:dcb} with a one-hot task embedding (best viewed in color).}
  \label{fig:dcb_detail}
  \vskip -0.2in
\end{figure*}

\myheading{DCB:}
An input image is resized to $\htimesw{320}{512}$ for computational efficiency (the original image is \htimesw{1050}{1680}), while the blurred image is obtained by applying a Gaussian filter on the original image with the standard deviation $\sigma = 2$. Both images are passed through a Panoptic-FPN with backbone network ResNet-50-FPN pretrained on COCO2017 \cite{kirillov2019panoptic}. The output of the Panoptic-FPN has 134 feature maps, consisting of 80 ``thing'' categories (objects) and 54 ``stuff'' categories (background) in COCO. Feature maps are then resized to
$\htimesw{20}{32}$ spatially, same as the discretization of fixation history. At a given time step $t$, feature maps $H$ for the original image and feature maps $L$ for the blurred image are combined for DCB:
\begin{equation}\label{eq:dcb}
    B_t = M_t\odot H + (1-M_t)\odot L
\end{equation}
where $\odot$ is element-wise product and $M_t$ is the mask generated from fixation history and repeated over feature channels (see \Fref{fig:dcb_detail}).
Note that the above equation is equivalent to  Eq. (1) in the main paper which is written in a recurrent form.

\myheading{Encoding the target object category:} The task embedding used in our model is the one-hot encoding maps which spatially repeat the one-hot vector. To make predictions conditioned on the task, inputs of each convolutional layer are concatenated with this embedding. This is equivalent to adding a task-dependent bias term for every convolutional layer.

\subsection{Implementation details}\label{sec:impl_details}
\myheading{Action Space}.
Our goal is to predict the pixel location where the person is looking in the image during visual search. To reduce the complexity of prediction, we discretize the image into a $\htimesw{20}{32}$ grid, with each patch corresponding to $\htimesw{16}{16}$ pixels in the original image coordinates. This descretized grid defines the action space for all models tested in this paper. At each step, the policy chooses one out of 640 patches and the center location of that selected patch in the original image coordinates is used as an action. The maximum approximation error due to this discretization procedure is $1.75$ degrees visual angle.

\myheading{IRL}.
The IRL model is composed of three components---the policy network, the critic network and the discriminator network. 
The \textbf{policy network} consists of four convolutional  layers whose kernel sizes are 5, 3, 3, 1 with padding 2, 1, 1, 0 and output channels are 128, 64, 32 and 1, and a softmax layer to output a final probability map.
The \textbf{critic network} has two convolutional layers of kernel size 3 and two fully-connected (fc) layers whose output sizes are 64 and 1. The convolutional layers have output sizes 128 and 256, respectively, and each is followed by a max pooling layer of kernel size 2 to compress the feature maps into a vector. Then this feature vector is regressed to predict the value of the state through two fc layers . The \textbf{discriminator network} shares the same structure with the IRL policy network except that the last layer is a sigmoid layer.
Note that all convolutional layers and fully-connected layers are followed by a ReLU layer except the output layer. 

\begin{figure*}[t!]
\begin{center}
\includegraphics[width=.9\textwidth]{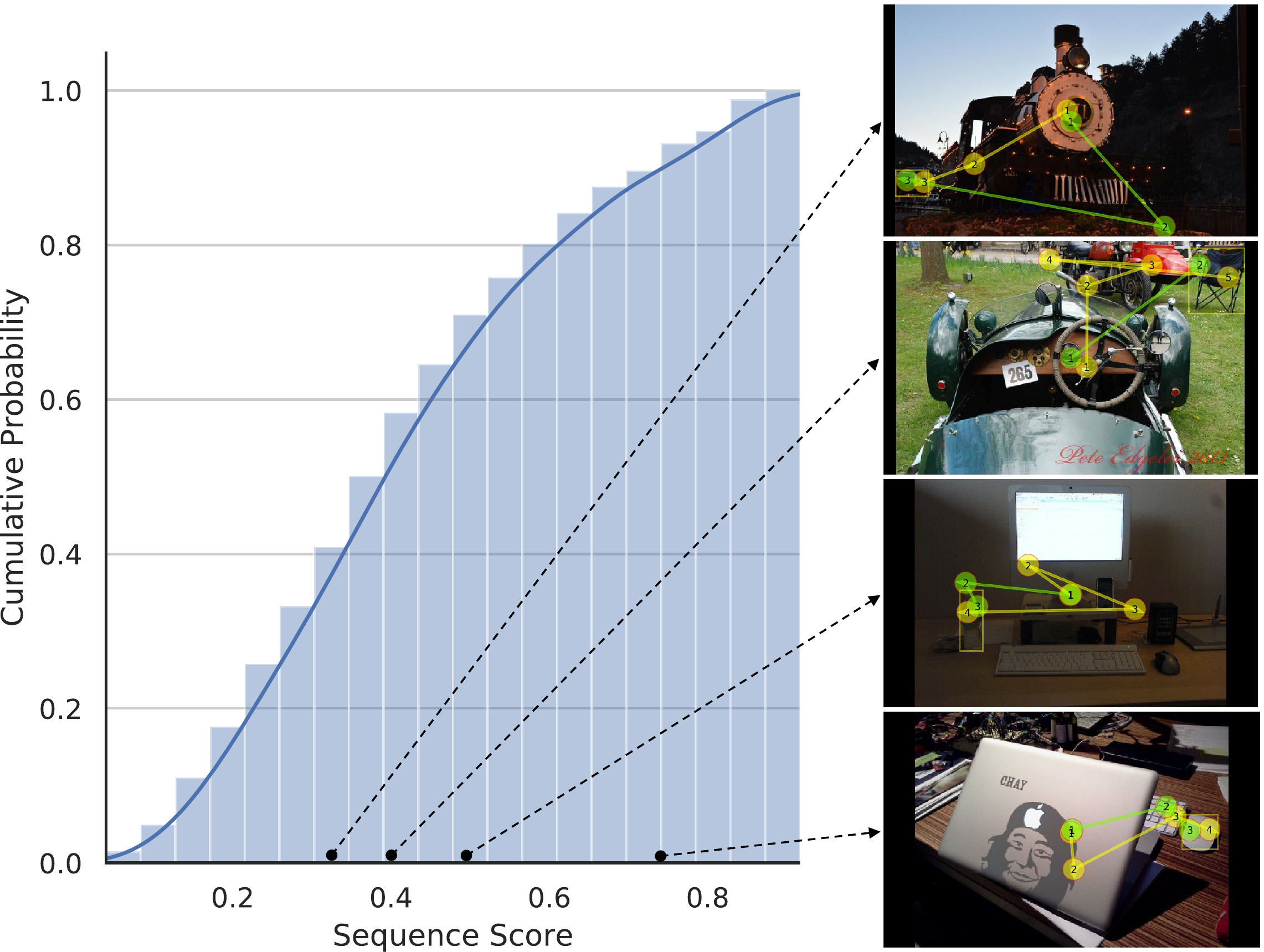}
\end{center}
\vskip -0.2in
  \caption{Left: cumulative distribution of the sequence scores of the proposed IRL scanpath prediction method. Right: Four qualitative examples. Human scanpaths are colored in yellow, and the IRL-generated scanpaths are in green. The sequence score for the generated scanpaths are 0.33, 0.40, 0.50, and 0.75, from top to bottom.}
\label{fig:seq_score_cdf}
\end{figure*}
The critic network is jointly trained with the policy network to estimate the value of a state (i.e., expected return) using smoothed $L_1$ loss. The estimated value is used to compute the advantage $A(S, a)$ (note that the state $S$ is represented by the proposed DCB in our approach) in Eq. (4) of the main paper using the Generalized Advantage Estimation (GAE) algorithm \cite{schulman2015high}. At each iteration, the policy network first generates two scanpaths by sampling fixations from the current policy outputs for each image in a batch. Second, we break the generated scanpaths into state-action pairs and sample the same number of state-action pairs from ground-truth human fixations to train the discriminator network which discriminates the generated fixation from behavioral fixations. Lastly, we update the policy and critic network jointly using the PPO algorithm \cite{schulman2017proximal} by maximizing the total expected rewards which are given by the discriminator (see Eq. (3) of the main paper).

\myheading{Training:}
The IRL model was trained for 20 epochs with an image batch size of 128. The batch sizes used for training the discriminator and policy networks were 64. For the PPO algorithm, the reward discount factor, the clip ratio and number of epochs were set to 0.99, 0.2, and 10, respectively. The extra discount factor in the GAE algorithm was set to 0.96. Both the policy network and the discriminator network were trained with a learning rate of  $0.0005$. It took approximately 40 minutes to train the proposed IRL model (for 20 epochs) on a single NVIDIA Tesla V100 GPU. The training procedure  consumed about 5.6GB GPU memory. Note that the segmentation maps used to construct the DCB state representation had been computed beforehand.

\myheading{Additional details on two baseline methods}. \textbf{Detector}: The detector network consists of a feature pyramid network (FPN) for feature extraction (1024 channels) with a ResNet50 pretrained on ImageNet as the backbone and a convolution layer for detection of 18 different targets. The detector network predicts a 2D spatial probability map of the target from the image input and is trained using the ground-truth location of the target. Another similar baseline is \textbf{Fixation Heuristics}.
This network shares exactly the same network architecture with the detector baseline but it is trained with behavioral fixations in the form of spatial fixation density map (FDM), which is generated from 10 subjects on the training images.

\myheading{Scanpath Generation}. 
When generating scanpaths, a fixation location is sampled from the probability map that the models have produced and Inhibition-of-Return is applied to prevent revisiting previously attended locations. All predictive methods including IRL, behavior cloning, and heuristic methods, generate a new spatial probability map at every step, while the predicted probability map is fixed over all steps for the Detector and Fixation Heuristic baselines.

\begin{table*}[t!]
\begin{center}
\begin{tabular}{l|c|c|c|c|cccc}
 \multirow{2}{2.2cm}{State Representation} & \multirow{2}{1.3cm}{\centering Sequence Score $\bm{\uparrow}$} & \multirow{2}{1.3cm}{\centering Scanpath Ratio $\bm{\uparrow}$} & \multirow{2}{1.2cm}{\centering TFP- AUC $\bm{\uparrow}$} & \multirow{2}{1.7cm}{\centering Probability Mismatch $\bm{\downarrow}$} & \multicolumn{4}{c}{MultiMatch $\bm{\uparrow}$} \\ 
 \cline{6-9}
 &  &  &  &  & shape & direction & length & position\\
\Xhline{1.2pt}
DCB & {\bf 0.422} & {\bf 0.826} & {\bf 4.509} & {\bf 0.987} & {\bf 0.886} & {\bf 0.695} & 0.866 & {\bf 0.885}\\
CFI & 0.402 & 0.619 & 3.412 & 1.797 & 0.875 & 0.666 & 0.864 & 0.857\\

\end{tabular}
\end{center}
\vskip -0.2in
\caption{{\bf Dynamic contextual belief (DCB) vs. cumulative foveated image (CFI)} under the framework of IRL.}
\label{table:cfi_cmpr}
\end{table*}

\begin{table*}[t!]
\begin{center}
\begin{tabular}{l|c|c|c|c|cccc}
 \multirow{2}{1.8cm}{State Representation} & \multirow{2}{1.3cm}{\centering Sequence Score $\bm{\uparrow}$} & \multirow{2}{1.3cm}{\centering Scanpath Ratio $\bm{\uparrow}$} & \multirow{2}{1.2cm}{\centering TFP- AUC $\bm{\uparrow}$} & \multirow{2}{1.7cm}{\centering Probability Mismatch $\bm{\downarrow}$} & \multicolumn{4}{c}{MultiMatch $\bm{\uparrow}$} \\ 
 \cline{6-9}
 &  &  &  &  & shape & direction & length & position\\
\Xhline{1.2pt}
DCB with all components & 0.422 & 0.803 & 4.423 & 1.029 & 0.880 & 0.676 & 0.841 & 0.888\\ 
w/o history map & 0.419 & 0.800 & 4.397 & 1.042 & 0.882 & 0.672 & 0.844 & 0.887\\
w/o saliency map & 0.419 & 0.795 & 4.403 & 1.029 & 0.880 & 0.675 & 0.840 & 0.887\\
w/o stuff maps & 0.407 & 0.777 & 4.111 & 1.248 & 0.876 & 0.662 & 0.836 & 0.875\\
w/o thing maps & 0.331 & 0.487 & 2.047 & 3.152 & 0.855 & 0.605 & 0.852 & 0.818 \\
w/o target map & 0.338 & 0.519 & 2.274 & 2.926 & 0.866 & 0.613 & 0.837 & 0.820  \\

\end{tabular}
\end{center}
\vskip -0.2in
\caption{{\bf Ablation study of the proposed state representation---dynamic contextual belief}. The full state consists of 1 history map, 1 saliency map, 54 stuff maps, 79 context maps and 1 target map. We mask out one part by setting the map(s) to zeros at each time.}
\label{table:abalation}
\end{table*}

\begin{table*}[t!]
\begin{center}
\begin{tabular}{l|c|c|c|c|cccc}
\multirow{2}{2.5cm}{Scanpath generation policy} & \multirow{2}{1.3cm}{\centering Sequence Score $\bm{\uparrow}$} & \multirow{2}{1.3cm}{\centering Scanpath Ratio $\bm{\uparrow}$} & \multirow{2}{1.2cm}{\centering TFP- AUC $\bm{\uparrow}$} & \multirow{2}{1.7cm}{\centering Probability Mismatch $\bm{\downarrow}$} & \multicolumn{4}{c}{MultiMatch $\bm{\uparrow}$} \\ 
 \cline{6-9}
 &  &  &  &  & shape & direction & length & position\\
\Xhline{1.2pt}
Based on total reward & {\bf 0.422} & {\bf 0.826} & {\bf 4.509} & {\bf 0.987} & {\bf 0.886} & {\bf 0.695} & 0.866 & {\bf 0.885}\\
Based on immediate reward & 0.375 & 0.704 & 3.893 & 2.143 & 0.886 & 0.648 & {\bf 0.873} & 0.852 \\

\end{tabular}
\end{center}
\vskip -0.2in
\caption{{\bf IRL model predictions using Greedy (immediate reward) and Non-greedy (total reward) policy}.}
\label{table:greedy_cmpr}
\end{table*}

\begin{table*}[t!]
\begin{center}
\begin{tabular}{l|c|c|c|c|cccc}
 & \multirow{2}{1.3cm}{\centering Sequence Score $\bm{\uparrow}$} & \multirow{2}{1.3cm}{\centering Scanpath Ratio $\bm{\uparrow}$} & \multirow{2}{1.2cm}{\centering TFP- AUC $\bm{\uparrow}$} & \multirow{2}{1.7cm}{\centering Probability Mismatch $\bm{\downarrow}$} & \multicolumn{4}{c}{MultiMatch $\bm{\uparrow}$} \\ 
 \cline{6-9}
 &  &  &  &  & shape & direction & length & position\\
\Xhline{1.2pt}
IRL, 20 ipc & 0.415 & 0.808 & 4.324 & 1.140 & 0.875 & 0.672 & 	0.832 & 0.879\\
CNN, 20 ipc & 0.408 & 0.723 & 3.906 & 1.325 & 0.884 & 0.664 &	0.849 & 0.878\\
\hline
IRL, 10 ipc & 0.409 & 0.774 & 4.029 & 1.318 & 0.881	& 0.591 & 	0.851 & 0.819\\
CNN, 10 ipc & 0.397 & 0.678 & 3.542 & 1.657 & 0.877	& 0.594 & 	0.847 & 0.821\\
\hline
IRL, 5 ipc & 0.389 & 0.723 & 3.696 & 1.603 & 0.876 & 0.588 & 0.844 & 0.813\\
CNN, 5 ipc & 0.388 & 0.678 & 3.484 & 1.731 & 0.886 & 0.594 &	0.862 &	0.828\\

\end{tabular}
\end{center}
\vskip -0.2in
\caption{{\bf Data efficiency of IRL and CNN}. ``ipc'' means images per category used for training. For exmaple, IRL 10 ipc means we train the IRL model using 10 images from each category which are randomly selected from the training data. CNN and IRL are trained and tested on the same images for fair comparison.}
\label{table:data_efficiency}
\end{table*}

\subsection{Additional Experiment Results}\label{sec:results}

\myheading{Cumulative distribution of sequence scores}. In the main paper we reported the {\it average} Sequence Score of 0.422 for the scanpaths generated by the IRL model. To put this in perspective, \Fref{fig:seq_score_cdf} plots the cumulative distribution of the sequence scores and shows four qualitative examples that have sequence scores of 0.33, 0.40, 0.50, and 0.75, respectively.

\myheading{Comparing different state representations}. To evaluate the benefits of having DCB as the state representation, we compared its predictive performance with the Cumulative Foveated Image (CFI)~\cite{zelinsky2019benchmarking} under the same IRL framework. CFI is created by extracting CNN feature maps on the retina-transformed images which are progressively more blurred based on the  distance away from the currently fixated location. On the other hand, the DCB is created by extracting panoptic segmentations~\cite{kirillov2019panopticfpn} on uniform-blur images which are uniformly blurred except around the fixated region (the level of blurriness applied in DCB is close to the middle level in the blur pyramid of CFI  \cite{zelinsky2019benchmarking,huang2015salicon,perry2002gaze}). For a fair comparison, we extract features for CFI using the backbone ResNet-50-FPN network from the Panoptic-FPN \cite{kirillov2019panopticfpn} that was used in DCB. Both DCB and CFI have the same spatial resolution of \htimesw{20}{32}. As shown in \Tref{table:cfi_cmpr}, the IRL model with DCB achieves significantly higher search efficiency and scanpath similarity than when using CFI as state representation. Specifically, DCB reduces the search gap by approximately 45\% and improves the scanpath ratio from 61.9\% to 82.6\%, much closer to the human behavioral ratio of 86.2\%. This result is even more impressive considering the size differences between the policy network used with DCB and CFI: DCB is trained with a smaller policy network, since it is comprised of 134 channels, nearly 8x smaller than CFI of 1024 channels. In our experiment, the policy network with CFI state representation has 29.6M parameters, while the policy network with DCB state representation only has 0.3M parameters. Relatedly, another benefit of having DCB as state representation is that it is memory and operation efficient. Creating DCB requires a smaller computational cost than creating CFI, since there's only a single level of blurriness in DCB and extracted panoptic segmentation maps are smaller by an order of magnitude than the feature maps extracted for CFI. Given that IRL models are particularly difficult to train in high dimensional environments~\cite{tucker2018inverse}, having an efficient representation like DCB can be very helpful.

\myheading{State Ablation}. 
DCB is a rich representation that incorporates top-down, bottom-up, and history information. The full representation consists of 136 belief maps, which can be divided into five groups: target object (1 map), ``thing'' (object, 79 maps), ``stuff'' (background classes, 54 maps), saliency (1 map, extracted using DeepGaze2 \cite{kummerer2017understanding}), and history (1 binary map for the locations of previous fixations). To understand the contribution of each factor, we removed the maps of each group one at a time and compared the resulting model's performance. As shown in \Tref{table:abalation}, target object and ``thing'' maps are the most critical for generating human-like scanpaths, followed by ``stuff'' maps, whereas saliency and history do not have strong impact to the model performance.


\myheading{Greedy vs. Non-greedy search behavior}.
How does human search behavior compare to generated scanpaths reflecting either Greedy or Non-greedy reward policies?
Under the greedy policy, the selection of each location to fixate during search reflects a maximization of immediate reward. But the greedy policy is highly short-sighted -- it only seeks reward in the short term. Non-greedy reward seeks to maximize the total reward that would be acquired over the sequence of fixations comprising a scanpath. This policy therefore does not maximize reward in the near term, but rather allows more exploration that leads to higher total reward. As shown in \Tref{table:greedy_cmpr}, we generated greedy and non-greedy policies from our IRL model and compared their predictive performance on human scanpaths. The results show that 1) models using greedy vs. non-greedy policy produce different search behaviors, with the model using non-greedy policy generating more human-like scanpaths by all tested metrics. This is an interesting finding. Despite the high efficiency of human search in our study (1-2 sec), the search process was strategic in that the fixations maximized total reward, even over that short period of time.

\myheading{Data Efficiency}.
Table \ref{table:data_efficiency} shows the full results of IRL and BC-CNN given different numbers of training images across different metrics. Both use DCB as the state representation. The results are consistent with the results presented in the main paper and suggest that IRL is more data-efficient when compared to the CNN -- IRL achieved comparable or better results than the CNN using less training data.

\end{document}